\documentclass[11pt]{article}

\usepackage[T1]{fontenc}
\usepackage{amsmath}
\usepackage[default]{sourcesanspro}
\usepackage{newtxmath}
\usepackage{microtype}

\usepackage{arxiv}
\definecolor{wbteal}{HTML}{28B894}

\usepackage{booktabs}
\usepackage{colortbl}
\usepackage{graphicx}
\usepackage[font=small,labelfont=bf,labelsep=period]{caption}

\usepackage[numbers,sort&compress]{natbib}
\definecolor{wbdeep}{HTML}{147C63}
\usepackage[colorlinks=true,linkcolor=wbdeep,citecolor=wbdeep,urlcolor=wbdeep]{hyperref}
\hypersetup{
  pdftitle={Tencent WorkBuddy Bench: A Multi-Domain Coding-Agent Benchmark with Contamination-Resistant Task Construction},
  pdfauthor={Tencent WorkBuddy Bench Team}
}

\shorttitle{Tencent WorkBuddy Bench}
\preprinttag{Technical Report \textperiodcentered\ 2026}

\logo{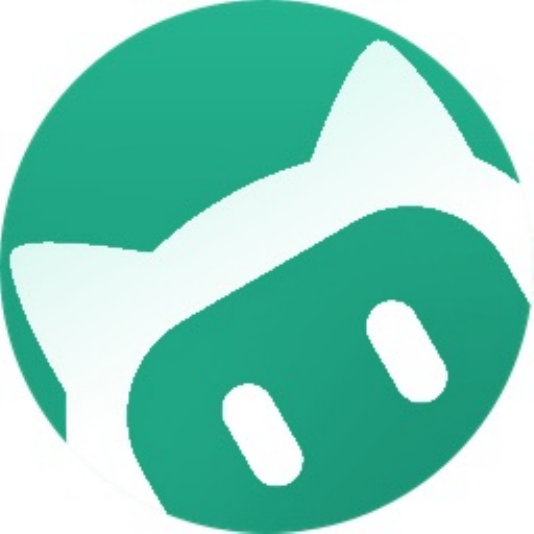}

\title{Tencent WorkBuddy Bench\\[0.34em]{\large A Multi-Domain Coding-Agent Benchmark with Contamination-Resistant Task Construction}}

\author{%
  {\large\bfseries Tencent WorkBuddy Bench Team}\\[0.35em]
  {\normalsize\color{black!70} Youtu Lab \(\cdot\) Keen Security Lab \(\cdot\) Workbuddy \(\cdot\) Yunding Security Lab}%
}
\date{\today}

\resourcelinks{%
  \faGlobe~\href{https://workbuddybench.com/}{Project page} \quad
  \faGithub~\href{https://github.com/Tencent/workbuddy-bench}{Code} \quad
  \faDatabase~\href{https://huggingface.co/datasets/tencent/workbuddy-bench}{Dataset} \quad
  \faFilePdf~\href{https://workbuddybench.com/report/main.pdf}{Report}%
}

\begin{document}

\maketitle

\begin{abstract}
In this paper we introduce \textbf{Tencent WorkBuddy Bench}, a multi-domain evaluation suite for
coding agents; this report documents its construction methodology, scoring protocol, and a
cross-model leaderboard. At its core is a unified evaluation framework for constructing and
running distribution-informed coding-agent tasks across four work domains -- Code,
Web, Office, and Security. Rather than adapting public issue text, every task is
reverse-engineered from a real commit, pull request, or business scenario and rewritten as a
short, colloquial, role-played request, so that a task's prompt is not recoverable by
web-searching the underlying issue, pull request, or commit thread. Because the dataset is
released openly -- task directories, environment images, evaluation harness, tests, and
reference solutions -- contamination resistance rests on this construction together with dataset
versioning rather than on secrecy. The four
subsets -- repository-level engineering, front-end development, office and business workflows, and
red-/blue-team security -- probe complementary facets of real work, each with its own verification
style. All are packaged in a uniform task-directory format and run, under a uniform and reproducible
protocol, on two agent harnesses (CodeBuddy Code and Claude Code); the full open release makes the
benchmark reproducible end to end and directly auditable, since any third party can re-run each
task and inspect its content. Because each subset uses a different scoring instrument, scores are not
comparable across subsets and the suite reports no suite-wide average. We report a cross-model
leaderboard across several model families.
\end{abstract}

\section{Introduction}

\begin{figure}[t]
\centering
\includegraphics[width=\textwidth]{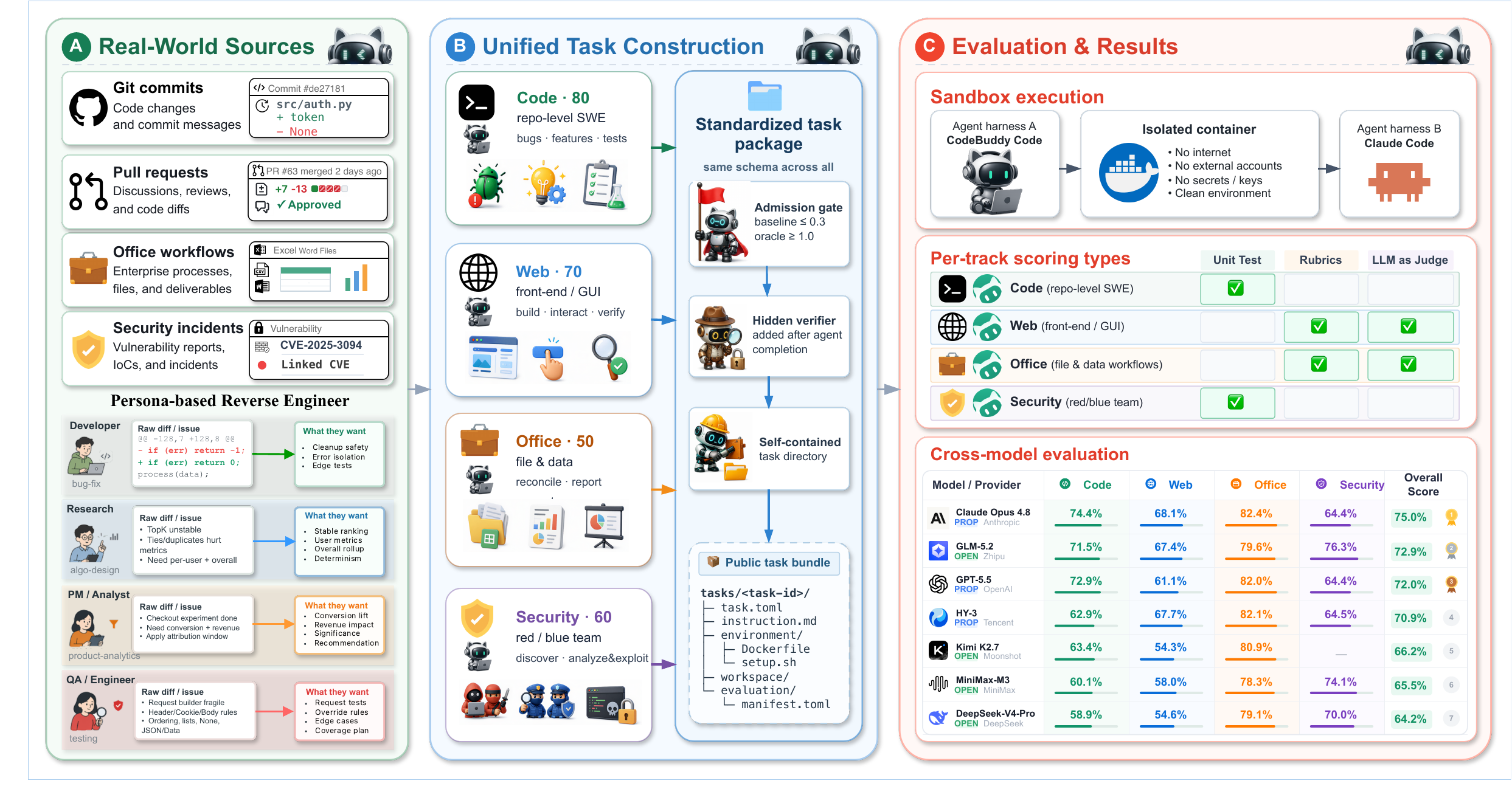}
\caption{Tencent WorkBuddy Bench at a glance. Real commits, pull requests, office workflows,
and security cases are reverse-engineered into colloquial, role-played requests, with task
distributions matched to real usage (left); the four tracks -- Code, Web, Office, and
Security -- share one open task-directory format (center); every task is scored in an isolated
sandbox under two agent harnesses (right).}
\label{fig:teaser}
\end{figure}

Coding agents are weighed today against two very different kinds of benchmark, each with a
different trade-off. Static, public suites such as SWE-bench and SWE-bench
Verified~\citep{jimenez2024swebench, openai2024sweverified} fix a task set at release time:
their problem statements, and often their solutions, circulate openly on the web, so a rising
score can reflect memorization of a specific issue thread or pull request rather than genuine
repository-level reasoning, and their scope is narrow -- overwhelmingly single-issue bug
resolution. The same crawlability problem holds beyond code: benchmarks for front-end
generation~\citep{si2024design2code} and web agents~\citep{zhou2024webarena} draw on public
repositories, screenshots, and websites that are themselves crawlable. Vendor production
benchmarks such as CursorBench~\citep{cursor2026cursorbench} take the opposite approach, drawing
tasks from real production sessions so that the task distribution tracks how an agent is
actually used -- but the benchmark itself is closed: an external party cannot inspect its task
distribution, rule out selection bias toward the vendor's own agent, or confirm that its task
mix generalizes beyond that vendor's user base. Evaluating agents meant to operate inside real
organizations therefore calls for a suite whose task distribution is informed by real work, that
resists the contamination path that matters most -- web-searchable prompts -- by construction
rather than only by novelty at release time, and that is released openly enough for an outside
party to re-run each task and audit its content directly.

We present \textbf{Tencent WorkBuddy Bench}, a multi-domain evaluation suite for coding agents;
this report documents its construction methodology, scoring protocol, and a cross-model
leaderboard. At its core is a unified evaluation framework for constructing and running
distribution-informed coding-agent tasks across four work domains -- \textbf{Code}, \textbf{Web},
\textbf{Office}, and \textbf{Security} (Figure~\ref{fig:teaser}) -- evaluated under a shared,
reproducible protocol on two agent harnesses (CodeBuddy Code and Claude Code). Code targets
repository-level software
engineering: locating, modifying, and verifying changes inside real open-source codebases under
role-played, colloquial requirements. Web targets front-end artifacts across generation,
modification, analysis, and quality assurance, from page implementation and data visualization to
stateful interaction, testing, reporting, and document conversion. Office targets
business workflows involving multiple files and deliverables. An agent must read mixed-format local
files, carry information across deliverables, update workspace state, and leave results that another
person can use. Its evaluation target is the final verifiable workspace state: deliverables, file
structure, state changes, evidence, and task-specific execution boundaries. Security spans the
security-team spectrum -- vulnerability discovery and safe reproduction, malware analysis, security
operations, and agent-security assessment -- rather than the writing of fixes. All four subsets
share a common task-directory format, a common admission protocol, and a common execution
infrastructure. What they do not share is a scoring instrument: Code uses hidden tests -- ``hidden'' meaning
held out from the agent while it solves, not withheld from the public, since the full test suite
ships in the open release -- Web a rubric with rule checks for deterministic constraints,
LLM/VLM judges for textual, structured, and visual semantics, and an agent-judge for interactive
state and workflow checks, Office a task-specific blend
of deterministic rule checks and semantic rubrics evaluated by an evidence-grounded LLM Judge, and
Security a deterministic scorer, so
scores are not comparable across subsets and the suite reports no suite-wide average -- a
deliberate design fact, not a gap to be closed. The unification is of construction and harness,
and the benchmark is released openly and is directly auditable: the protocol, task format, task
directories, environment images, evaluation harness, tests, and reference solutions are all
public, so any third party can re-run each task and inspect its content.

\textbf{Why these four belong in one suite.} A coding agent placed in real organizational work no
longer only edits code: the same agent is asked to build a web front-end, produce or reconcile an
office document, and reason about a security artifact. Code, Web, Office, and Security are the four
artifact and workflow boundaries that this work crosses, and the suite treats them as one because
the task shape is identical at every boundary -- the agent is dropped into a workspace, produces
an artifact from a natural-language request, and is graded by a verifier it never sees. That
shared shape, not a shared scoring rule, is what makes the four subsets one suite.

Resistance to the contamination path that matters most -- web-searchable prompts -- is a
first-class design constraint, not an afterthought. Tasks are not
reproductions of public issue titles or tutorial exercises: each is reverse-engineered from a
real commit, pull request, or business scenario and rewritten as a short, colloquial,
role-played request whose instruction withholds the root cause, the reference diff, and any
framing that would hand the agent the solution, so a task's prompt is not recoverable by
web-searching the underlying issue or pull-request thread. Because the dataset is released
openly, that construction-level resistance is backed by dataset versioning rather than by
secrecy; Section~\ref{sec:benchmark-design} details the mechanism and scopes honestly what it
does and does not resist.

Task distributions are informed by analysis of real usage, not by reuse of real usage data.
Each subset's mix of categories, task modes, and difficulty is matched against internal usage
taxonomies -- query-intent categories and request-structure patterns -- so that, for example,
Code's 80 tasks span five requester roles and task types well beyond bug fixing
(Section~\ref{sec:the-benchmark}). What is analyzed and matched is the \emph{distribution} of
real requests, not the requests themselves: no raw user prompt, session, or user data is reused
or exposed in a released task. It is also what lets the suite be released in full and
audited openly, where raw-session benchmarks face privacy constraints that limit
disclosure.

\textbf{Contributions.} This report makes four contributions.
\textbf{1)}~We introduce Tencent WorkBuddy Bench, a suite of four parallel subsets -- Code, Web, Office,
and Security -- that evaluate coding agents on repository-level software engineering, front-end
web development, office and business workflows, and security-team workflows under one
reproducible harness.
\textbf{2)}~We construct tasks with a distribution-informed methodology that resists prompt
contamination: each is reverse-engineered from a real commit, pull request, or business
scenario, matched against internal usage taxonomies, and rewritten as a colloquial, role-played
natural-language requirement, not reproduced from public issue text or drawn from user sessions.
\textbf{3)}~We develop an evaluation methodology that reaches beyond pass/fail unit tests. Every
admitted task clears baseline/oracle admission gates (baseline reward $\le 0.3$, oracle reward
$\ge 1.0$), confirming that the untouched workspace does not already pass and that at least one
reference solution reaches full verifier reward; front-end artifacts are scored through rule checks for
deterministic constraints, LLM/VLM judges for textual, structured, and visual semantics, and an
agent-judge that drives the running artifact to inspect interactive flows and state. For Office,
deterministic rule checks verify files, structure, values, state,
and execution boundaries, while an LLM Judge evaluates binary semantic rubrics using fixed
evidence extracted after the task ends. The two scores are reported separately and combined using
each task's preconfigured weight.
\textbf{4)}~We report a cross-model leaderboard spanning multiple model families under two
evaluation harnesses (CodeBuddy Code and Claude Code).

In short, this report provides three things: the suite's design, the current task-set
composition of each subset, and the cross-model leaderboard. Table~\ref{tab:suite} summarizes
the four subsets.

\begin{table}[t]
\centering
\caption{Suite at a glance: the four subsets, each scored under a dual harness (CodeBuddy Code
and Claude Code). All subsets are the initial public release.}
\label{tab:suite}
\small
\begin{tabular}{@{}l l l p{5.6cm}@{}}
\toprule
Subset & Domain & Scale & Metric \\
\midrule
Code & Repository-level SWE & 80 tasks & Hidden-test score per run \\
Web & Front-end / GUI & 70 tasks & Rubric scoring (rule / LLM-VLM / agent) \\
Office & Office data \& file workflows & 50 tasks & Task-specific Rule/Judge blend \\
Security & Red- \& blue-team security & 60 tasks & Programmatic \texttt{scoring.py} (no LLM judge) \\
\bottomrule
\end{tabular}
\end{table}

The remainder of the report is organized as follows. Section~\ref{sec:related-work} first
positions the suite against existing public and vendor-production agent benchmarks in the code
and web domains.
Section~\ref{sec:benchmark-design} then describes the suite's shared design principles, task
format, and execution model, followed by sections detailing each of the four subsets, the
evaluation harness and scoring methodology, results, and limitations.

\section{Task Construction}
\label{sec:benchmark-design}

This section states the suite-level construction protocol of Tencent WorkBuddy Bench. Across all
four subsets -- Code, Web, Office, and Security -- tasks follow the same broad stages: sourcing,
rewriting into realistic requests, assembling the agent-visible workspace, isolating evaluation
assets until the episode ends, and packaging each task as a self-contained directory. User data is
kept out of construction throughout. The scoring instruments and any subset-specific admission
checks are not uniform: Code uses hidden tests, Web uses rule, LLM/VLM, and agent-judge rubric
items, Office uses a task-specific blend of deterministic rule checks and evidence-grounded LLM
Judge rubrics, and Security uses a deterministic \texttt{scoring.py}.
Because these instruments differ, scores are not comparable across subsets and the suite reports
no suite-wide average; this is a deliberate design decision, not a limitation to be corrected.
This section covers construction and packaging; execution and per-track scoring are specified
once, in Section~\ref{sec:harness}.

\paragraph{Task sources.} Every task is anchored to a concrete origin of one of two kinds: a real
upstream artifact -- a historical commit or pull request in an open-source repository (Code), or
a real, historical CVE (Security's whitebox-audit tasks) -- or a concrete business scenario
(Web, Office, and the synthetic task families of Code and Security). Which scenarios are worth
building, and in what proportion, is decided against internal usage taxonomies: each subset's mix
of categories, task modes, roles, and difficulty is matched to the \emph{distribution} of real
requests, never to the requests themselves. No raw user prompt, session, or other user data
enters any task; construction is informed only by aggregate distributions. Within the
business-scenario branch, Office uses two construction routes: tasks reconstructed from task
specifications and target capabilities, and tasks expanded from abstracted office workflows. Both
are packaged as self-contained workspaces containing only openly shareable inputs and pass the same
checks for workspace integrity, evaluation assets, calibration, and release readiness.

\paragraph{Rewriting protocol.} Tasks are not written as tidy issue titles or textbook exercises.
Where a task derives from a real upstream artifact, its original context is reverse engineered
and rewritten as a short, colloquial, underspecified natural-language request; where it is
authored from a business scenario, the request is written directly in the same voice. Either way
the request reads as a plausible ask from a colleague or customer -- Code additionally voices
every task through one of five requester personas (developer, algorithm engineer, product
manager, QA, operations), and Security assigns each task domain a professional role -- and the
instruction withholds the root cause, the reference diff, and any framing that would hand the
agent the solution, so the agent must locate the relevant surface of the workspace itself before
it can act. This stands in contrast to benchmarks that supply a detailed, already-diagnosed issue
report.

\paragraph{Deliberate underspecification.} Across all four subsets, requests are written the way a
colleague actually asks -- an intent and a constraint, not a specification -- and are deliberately
left underspecified: they routinely omit the target file or module, the exact schema or interface,
edge-case handling, and the precise boundary of the change. Resolving these gaps is part of the
task itself: the agent must recover the missing context from the workspace -- the repository, data
fixtures, or existing code and interfaces -- and commit to a reasonable implicit assumption rather
than being handed one. This is intentional, not an oversight: it tests requirement disambiguation
and grounding as much as code synthesis, and is what separates a realistic work request from a
tidy issue title. Reward is computed from task-specific checks, rubrics, or evaluation procedures,
not by matching one reference implementation or phrasing. These instruments encode the intended
contract, so a plausible but contract-violating output can still fail. The agent is judged on
meeting that contract, not on recovering one blessed realization of it.

\paragraph{Post-episode evaluation isolation.} Throughout an episode, the agent sees the task
instruction and declared workspace but not the grading assets. Only after the agent has finished
acting are task-specific checks, rubrics, or evaluation procedures introduced into the sandbox or
invoked by the evaluation pipeline. ``Hidden'' or ``held out'' therefore describes solve-time
visibility, not secrecy after release: the evaluation assets are public with the rest of the
dataset. Their form remains subset-specific -- Code hidden tests, Web rubric evaluators, Office
Rule--Judge evaluation, and Security's deterministic scorer -- while the shared property is the
temporal boundary between acting and grading. Code's diagnostic gold patch and oracle-gated
admission are described in the Code subsection of Section~\ref{sec:the-benchmark}.

\paragraph{Task-directory format.} Tasks are packaged using a Harbor~\citep{harbor}-style task-directory
convention, with a small delta from the vanilla Harbor layout that separates the agent-visible
workspace from post-episode evaluation assets:

\begin{codeblock}
\noindent
tasks/<task-name>/\\
\hspace*{1.5em}task.toml\\
\hspace*{1.5em}instruction.md\\
\hspace*{1.5em}environment/\\
\hspace*{3em}Dockerfile\\
\hspace*{3em}workspace/\\
\hspace*{1.5em}tests/\\
\hspace*{3em}test.sh\\
\hspace*{3em}grading/\\
\hspace*{3em}gold.patch \ (optional; Code diagnostic reference)
\end{codeblock}

\texttt{instruction.md} carries the natural-language request described above. \texttt{task.toml}
carries task metadata -- category, difficulty, tags, resource limits, and per-role timeouts --
under a versioned schema. \texttt{environment/} defines the sandbox: a Dockerfile that copies in
\texttt{workspace/} and nothing else, so that the agent-visible surface is exactly the repository
or business artifact under test. \texttt{tests/} holds post-episode evaluation assets:
\texttt{test.sh} is the entry point and \texttt{grading/} contains the task-specific checks or
evaluation configuration. The optional \texttt{gold.patch} is a Code-specific diagnostic
reference; its role in Code's oracle-gated admission is described in
Section~\ref{sec:the-benchmark}. Because the Dockerfile builds only the visible workspace and
everything under \texttt{tests/} stays outside the image, the post-episode evaluation boundary is
a property of the packaging itself, not of runtime configuration.

\paragraph{Execution.} How a packaged task is executed -- the sandboxed container lifecycle,
model connectivity, harness backends, and each track's scoring rule -- is specified once, in
Section~\ref{sec:harness}.

\paragraph{Contamination-resistant task construction.} Resistance to contamination comes first
from \emph{construction}. Because tasks are built from real commits, CVEs, and business scenarios
and rewritten into role-played natural-language requests rather than copied from public problem
statements, no task's instruction text is recoverable by web-searching the underlying issue,
pull-request, or commit thread: the searchable-prompt path is closed at the point the task is
written, independent of when the task is released. Because the dataset is released openly -- task
directories, grading tests, and reference solutions included -- this resistance can no longer lean
on secrecy or on withholding the graded answer. Two open-benchmark mechanisms carry the remaining
weight: dataset \emph{versioning}, under which the suite is periodically refreshed and
re-versioned so that a released snapshot can be superseded once exposure to it accumulates, and
optional canary strings that let a later training crawl of the released set be detected. The
residual exposure is stated in the same breath: a model may already have seen the original public
commit or pull-request code, or -- for the CVE-anchored security tasks -- public vulnerability
analysis of the underlying flaw; and, exactly as for any openly released benchmark such as
SWE-bench, a public task set is subject to post-release training exposure that versioning
mitigates but does not eliminate. The claim is therefore narrow and honest: contamination-resistant
\emph{task construction} closes the searchable-prompt path by construction, and open-release
versioning manages exposure over time -- not that the suite is contamination-free.

\paragraph{Version naming.} Each subset carries an internal version identifier combining a major
index with a date stamp, but the semantics of the major index are subset-specific rather than
uniformly sequential, so version numbers are not directly comparable across subsets.

\subsection{What Is Released}
\label{sec:disclosure}

The suite is released as a fully open, SWE-bench-style dataset: everything needed to run, grade,
and audit a task is public. The construction protocol above, the packaging convention, the task
directories, the workspace/environment images, the evaluation harness and its aggregation code,
the grading tests, and the reference solutions are all released, so any third party can re-run an
individual task and inspect its content directly -- the benchmark is fully reproducible and openly
auditable, not merely auditable at the level of a published protocol. Table~\ref{tab:disclosure}
lists what the release contains. The one thing it does not contain is user data, and that is by
absence rather than by withholding: no raw user prompt, session, or other user data is used at any
point in construction, so there is none to release.

\begin{table}[h]
\centering
\caption{Open release: the dataset ships every component needed to run, grade, and audit a task.}
\label{tab:disclosure}
\begin{tabular}{@{}p{9.1cm} p{4.3cm}@{}}
\toprule
Component & Status \\
\midrule
Task-directory skeleton and packaging convention (this section) & Released \\
Task prompts and instruction text & Released \\
Workspace/environment images for offline third-party testing & Released \\
Evaluation harness and score-aggregation code & Released \\
Grading tests (the verifier held out from the agent at solve time) & Released \\
Gold patches and reference solutions & Released \\
Per-task and aggregate scores, and the public leaderboard & Released \\
\midrule
User data of any kind & Not used at all -- none exists to release \\
\bottomrule
\end{tabular}
\end{table}

\section{The Benchmark}
\label{sec:the-benchmark}

Tencent WorkBuddy Bench is organized into four complementary subsets -- Code, Web, Office, and
Security -- each targeting a distinct class of realistic agentic tasks while sharing a common
task format and scoring philosophy. This section introduces the Code subset; the following
sections cover Web, Office, and Security in turn.

\subsection{Code}

\begin{figure}[t]
\centering
\includegraphics[width=\textwidth]{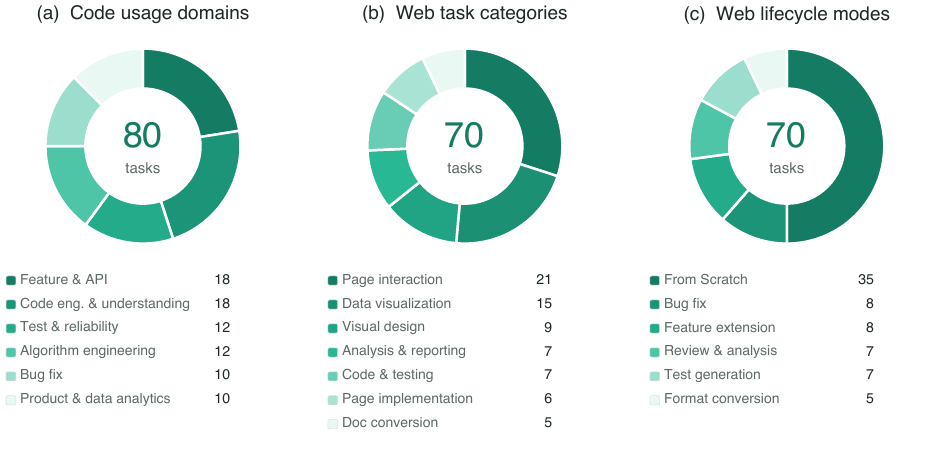}
\caption{Code and Web task composition. \textbf{(a)} Six Code usage domains merged from 18
fine-grained categories; bug fixing accounts for 10 of 80 tasks. \textbf{(b)} Seven Web task
categories. \textbf{(c)} Six Web lifecycle modes; From Scratch accounts for 35 of 70 tasks.}
\label{fig:code-web-composition}
\end{figure}

The Code subset measures whether an agent can carry out a real, role-played engineering request
against a full open-source repository -- not a single-file toy problem, and not a bug report
handed to it pre-diagnosed. The agent is dropped into a project checked out at a baseline commit,
must locate the relevant code across modules, make the change, and keep the project's hidden
tests green. What sets the subset apart is its role and task-type diversity: every task is voiced
by one of five requester roles -- developer, algorithm engineer (algo), product manager (pm),
quality assurance (qa), and operations (ops) -- and spans far more than bug-fix work
(Table~\ref{tab:code-composition}, Figure~\ref{fig:code-web-composition}(a)).

\paragraph{Task provenance.} Each task expresses its target change as a natural-language,
role-played request, so solving it requires reading and reasoning about the repository itself.
Of the 80 tasks, 34 are
anchored to a real upstream commit against an actual OSS snapshot (Family~A); the remaining 46
have no upstream code and divide -- with an approximate internal split -- between clean-room
reimplementations (Family~B, 24 tasks, including the 4 tasks that port
JavaScript/TypeScript/Rust targets into Python) and fully synthetic workspaces (Family~C, 22
tasks), as summarized in Table~\ref{tab:code-provenance}. Published repository counts vary with
whether clean-room and ported targets are included, so we do not report an aggregate count.

\begin{table}[h]
\centering
\caption{Code subset provenance families. Counts sum to the 80-task release: A = 34, B + C = 46.}
\label{tab:code-provenance}
\begin{tabular}{@{}l p{6.0cm} c p{4.6cm}@{}}
\toprule
Family & Definition & Count & Example \\
\midrule
A & Real OSS snapshot at an upstream commit; the gold patch is the actual human fix & 34 &
Django, Flask, pytest, Black, Pydantic, httpx, Celery ($\sim$18 repositories) \\
B & Clean-room \texttt{*\_like} reimplementation of a target library's public API, no original
code copied; includes the 4 cross-language ports (JS/TS/Rust originals in Python) & 24 &
\texttt{fastapi\_like/openapi.py} stub rather than FastAPI itself \\
C & Fully synthetic workspace with CSV/JSON fixtures, authored to exercise a role's workflow
directly & 22 & algo workspaces (12) and pm data workspaces (10) \\
\bottomrule
\end{tabular}
\end{table}

\paragraph{Scale and release.} Code comprises 80 tasks. Each ships as a self-contained
Harbor-style task directory -- \texttt{instruction.md}, \texttt{task.toml} metadata, an
\texttt{environment/} Docker snapshot of the target repository, and a \texttt{tests/} directory
holding hidden tests plus a diagnostic \texttt{gold.patch} -- following the task-directory format
of Section~\ref{sec:benchmark-design}.

\paragraph{Oracle-gated admission.} Each candidate Code task passes a two-run validation before
admission. The task image is first built and its verifier is run against the unchanged baseline
workspace. The task's \texttt{solution/solve.sh} then applies the diagnostic gold patch, after which
the verifier is run again. Admission requires baseline reward $\leq 0.3$ and oracle reward $=1.0$.
This removes tasks whose initial workspace already satisfies too much of the intended contract, as
well as tasks whose gold patch cannot achieve full verifier reward. The gold patch is a diagnostic
reference for this validation, not the unique correct solution; any patch that satisfies the
hidden tests receives the corresponding reward.

\begin{table}[h]
\centering
\caption{Code subset composition (80-task open release).}
\label{tab:code-composition}
\begin{tabular}{@{}l p{9.5cm}@{}}
\toprule
Dimension & Breakdown \\
\midrule
Roles & developer 30 \(\cdot\) algo 19 \(\cdot\) pm 15 \(\cdot\) ops 10 \(\cdot\) qa 6 \\
Difficulty (editorial) & easy 7 \(\cdot\) medium 31 \(\cdot\) hard 42 \\
Difficulty (L-ladder) & L2 4 \(\cdot\) L3 27 \(\cdot\) L4 40 \(\cdot\) L5 9 (centered on L4) \\
Admission gate & baseline \(\leq\) 0.3, oracle (gold patch) \(=\) 1.0 against hidden tests \\
\bottomrule
\end{tabular}
\end{table}

\paragraph{Domains and difficulty.} Tasks carry one of 18 fine-grained categories, merged into
six usage domains for readability (Figure~\ref{fig:code-web-composition}(a)). Bug fixing accounts for only
10 of the 80 tasks; the other five domains -- feature and interface work, code engineering,
testing, algorithm engineering, and product/data analytics -- carry the remaining 70, a
deliberate expansion beyond the ``fix a bug, add a feature'' framing of earlier benchmarks.
Difficulty comes chiefly from cross-module exploration -- finding \emph{where} to edit rather
than \emph{how} -- and grows with repository size and structure as tasks climb the L-ladder, a
repository-complexity scale running from L2 (small, few modules) to L5 (large multi-module
codebases). Table~\ref{tab:code-composition} gives the role and difficulty distributions.

Early evaluation runs during construction showed what failure looks like at repository scale: the
dominant zero-score modes were agents looping on test-file edits until timeout, and agents losing
their way in a large codebase and editing entirely the wrong files -- evidence that difficulty
comes from navigation and grounding rather than code synthesis.

A representative role-played request (product manager, product-analytics,
\texttt{signup\_funnel}, hard):

\begin{quote}
\itshape
``The checkout-copy experiment finished; I want to know first whether the new version is
better. The data has impression and purchase events -- please compute per-group conversion,
revenue, and a simple conclusion, and don't count purchases that happen long afterward.''
\end{quote}

The request states an intent and a constraint, not an implementation plan: it names neither the
relevant file, the expected schema, nor how the attribution window for excluding late purchases
should be drawn, leaving the agent to recover that context from the repository itself.

\begin{figure}[t]
\centering
\includegraphics[width=\textwidth]{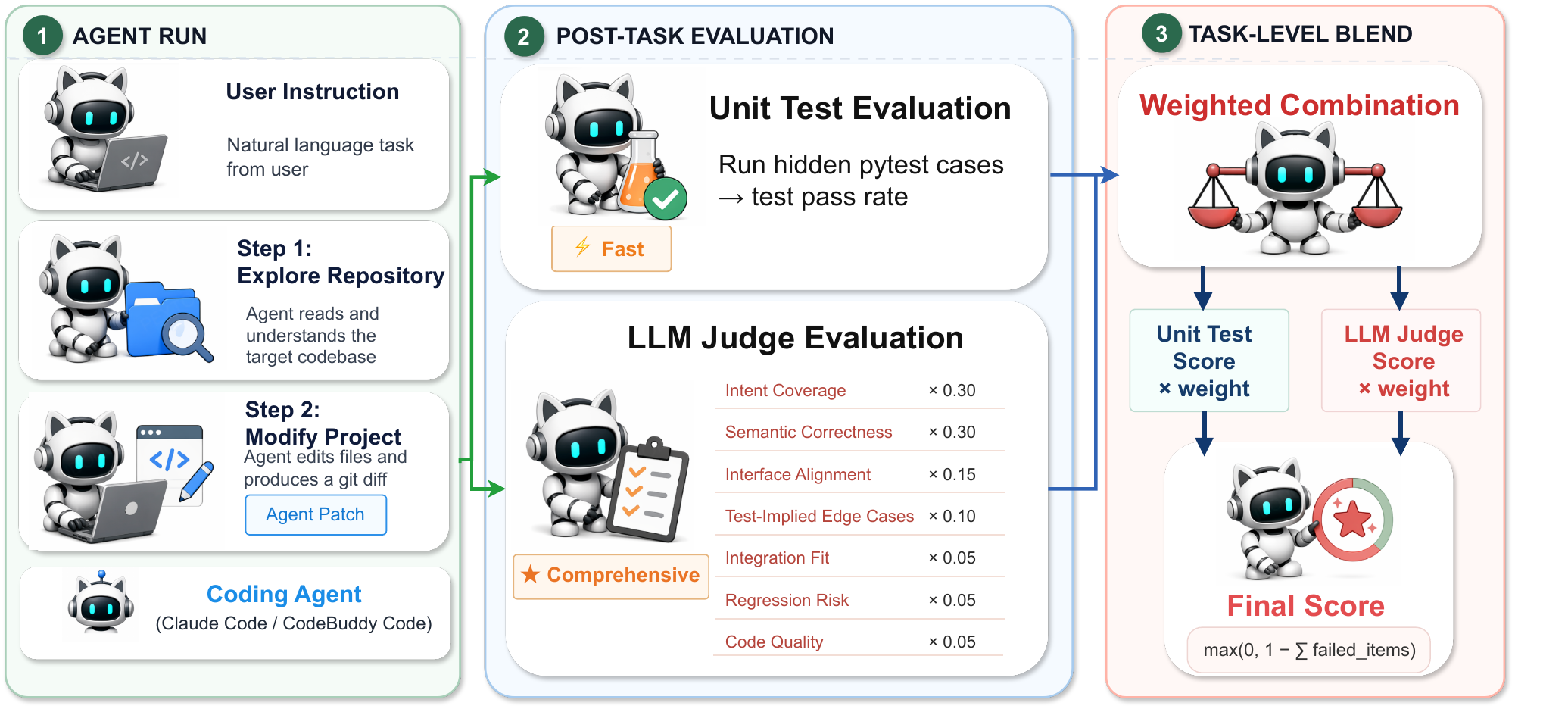}
\caption{Code task and evaluation workflow. The coding agent reads a natural-language request,
explores the repository, and emits a patch (left); the patch is graded by hidden unit tests and,
as a diagnostic reference, by a rubric-weighted LLM judge (center). The headline Code metric is the
hidden-test score; the LLM-judge reading and its blend are reported as reference values only and
never enter the headline metric (right).}
\label{fig:code-workflow}
\end{figure}

\paragraph{Scoring.} Each task is scored by a per-task verifier run inside its Docker image after
the agent's patch is applied; the headline Code metric is the \emph{run-level score}, the per-run
average of per-task hidden-test scores. Section~\ref{sec:harness} gives the three verifier forms,
gold-patch handling, and the reference readings in full. Figure~\ref{fig:code-workflow} summarizes
this task and evaluation workflow.

\subsection{Web}

The Web subset tests whether a model can deliver a runnable, checkable front-end artifact -- not simply
emit plausible-looking HTML in a chat turn. Every task carries an artifact-not-chat contract: the
agent must produce a runnable artifact at a declared output path (for example, an HTML entry
point); a well-written answer with no artifact at that path fails regardless of content. Across
its 70 tasks, coverage spans front-end artifact generation, modification, analysis, and quality
assurance in one task space: page implementation, page interaction, data visualization, visual
design, analytical reporting, code testing, and document conversion.

Tasks are organized into seven categories (Figure~\ref{fig:code-web-composition}(b)): page interaction
(21 tasks) and data visualization (15) dominate, together accounting for 36 of the 70 tasks,
while the remainder covers visual design, front-end project analysis, code testing, page
implementation, and document conversion -- work that traditional front-end generation benchmarks
rarely exercise.

Orthogonally, each task is authored to exercise one point in the web-development lifecycle
(Figure~\ref{fig:code-web-composition}(c)). From Scratch alone would only probe generation ability, so half the
suite instead requires fixing front-end state, runtime, or visual defects, extending an existing
page or application, reviewing Web project evidence, generating regression tests, or converting
source material into a front-end-facing deliverable -- so that models which can only create, and
not maintain, are not rewarded disproportionately.
In panel (c), From Scratch holds exactly half the suite (35 of 70 tasks), with the other half
split across bug fix (8), feature extension (8), review \& analysis (7), test generation (7), and
format conversion (5).

A third axis tracks interaction and state complexity. Twenty-five tasks are noninteractive
front-end project artifacts, while 45 require interaction or state: single-flow state changes
(15), persistence, offline, or cross-state behavior (13), multi-step workflows (9), and light
interaction (8). This axis keeps the subset from collapsing into static page generation: many
tasks require the artifact's state to change, recover, or stay consistent under user actions.

A representative request from the page-interaction category (mobile store booking):

\begin{quote}
\itshape
``I want a mobile store-booking page: users pick a service and a time slot, fill in contact
details, and confirm. Full slots must not be selectable, and there should be a review step
before submitting.''
\end{quote}

The ask names an intent and a couple of constraints -- slot capacity, a review step before
submission -- not a full specification, leaving the agent to produce a runnable artifact that a
rubric can verify against the requested behavior.

\begin{figure}[t]
\centering
\includegraphics[width=\textwidth]{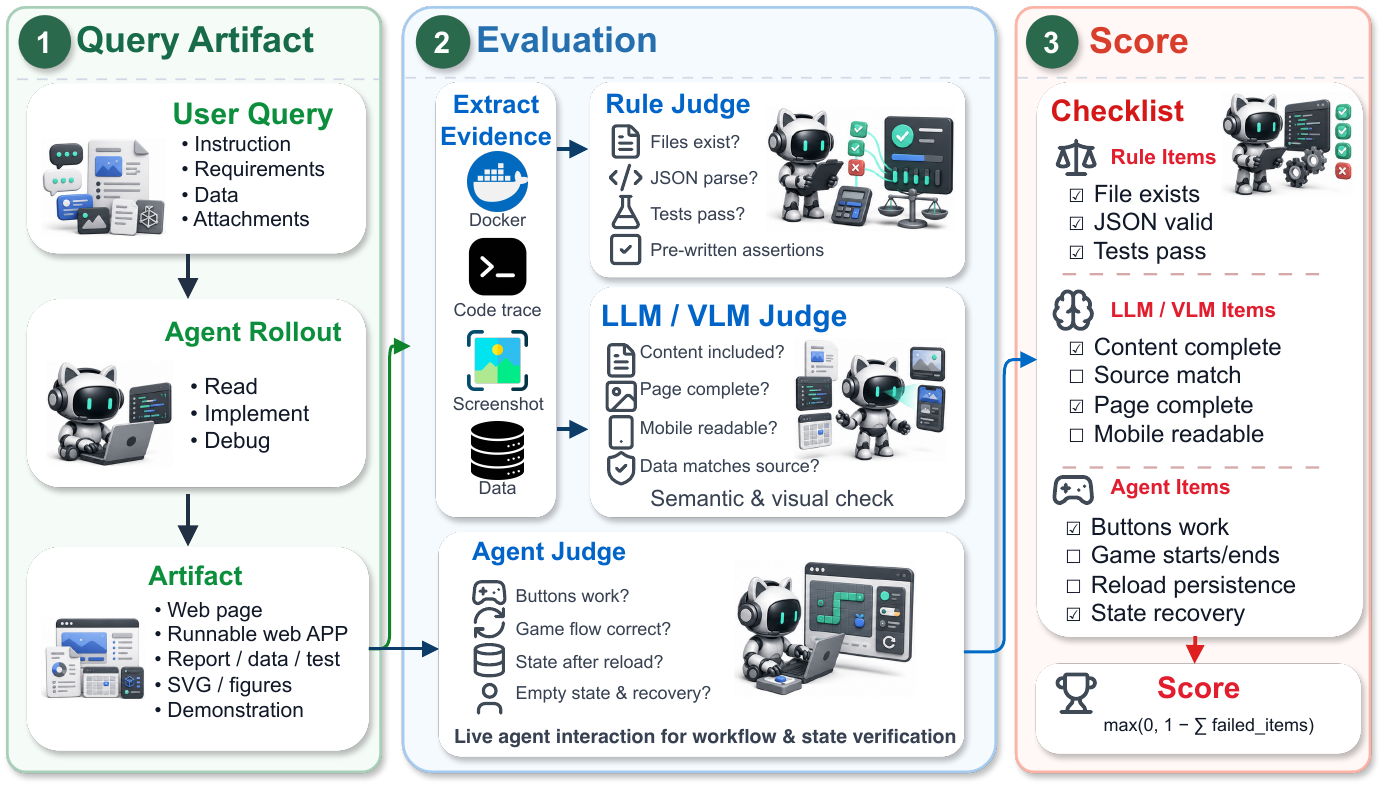}
\caption{Web task and evaluation workflow, from query and agent rollout to the delivered artifact
(left), through rule, LLM/VLM, and agent judges over extracted evidence (center), to rubric-item
checklist scoring (right), combining deterministic checks, semantic and visual judgment, and live
interaction over the delivered artifact.}
\label{fig:web-workflow}
\end{figure}

Figure~\ref{fig:web-workflow} summarizes this artifact-centered workflow, from query interpretation
and agent rollout to evidence extraction, complementary judges, and checklist scoring.

Scoring uses rubric items judged by rule checks, LLM/VLM judges, and an agent-judge. Rule checks
cover deterministic delivery constraints such as files, formats, prechecks, and executable tests;
LLM/VLM judges review textual, structured, DOM, screenshot, and visual evidence; and the
agent-judge drives the running artifact to inspect workflows, state changes, and persistence. A
run must still deliver the declared artifact at the declared output path, and tasks run with no
access to the live internet, external accounts, keys, or live data. Section~\ref{sec:harness}
gives the item counts, aggregation rule, and model-judge risk in full.

\subsection{Office}
\label{sec:subset-office}

The Office subset tests whether an agent can complete a natural-language work request in a local
workspace containing mixed-format files. Inputs include spreadsheets, documents, PDFs, JSON exports,
Markdown notes, and file trees; outputs include updated workbooks, reports, structured records, state
files, and handoff material. The agent must produce the requested deliverables, keep information
consistent across files, update related state, preserve evidence for review, and respect task-specific
execution constraints. Evaluation examines the final workspace, which catches failures that a
text-answer score misses, such as writing a plausible summary without updating the workbook it
describes or creating a file while leaving dependent state inconsistent.

\begin{figure}[t]
\centering
\includegraphics[width=\textwidth]{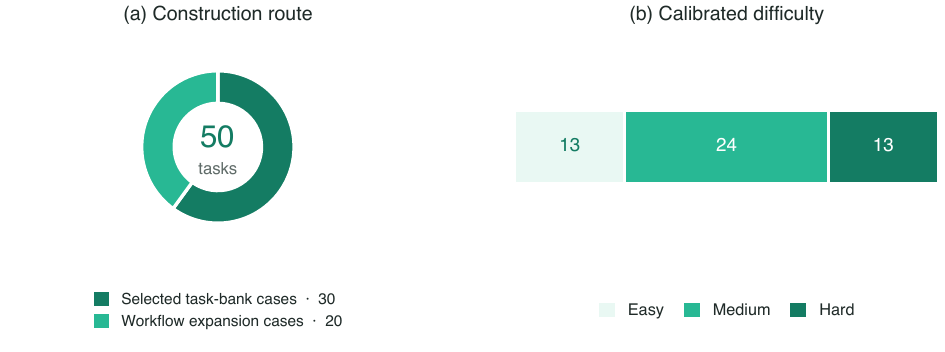}
\caption{Composition of the 50-task Office release by construction route and calibrated difficulty.}
\label{fig:office-composition}
\end{figure}

\begin{figure}[t]
\centering
\includegraphics[width=\textwidth]{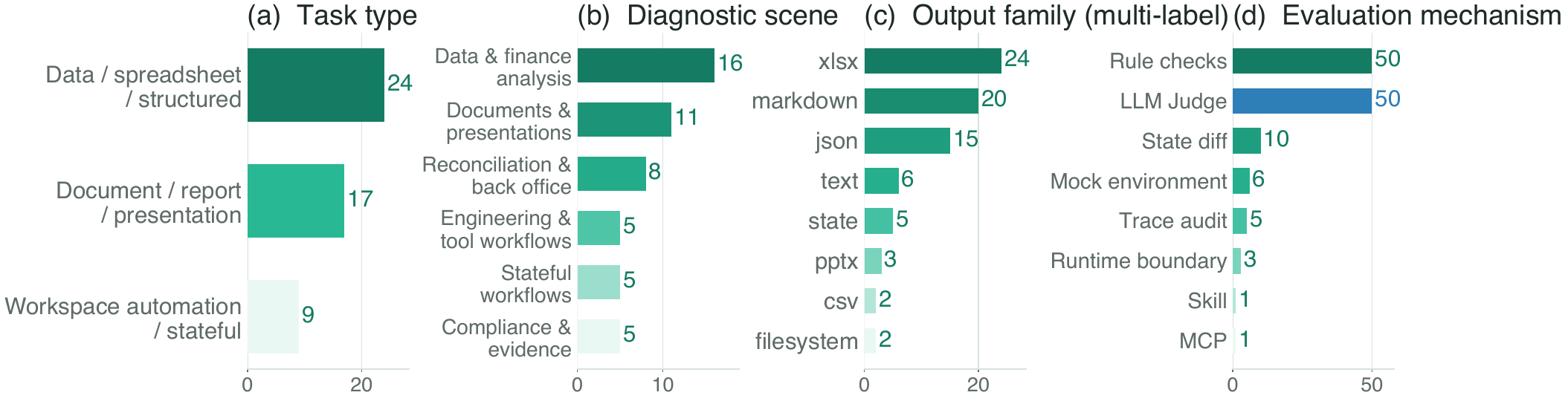}
\caption{Office coverage across task type, diagnostic scenario, output family, and evaluation
mechanism, arranged as a single row of four bar charts. Output families and mechanisms use
multi-label counts; scenario groups describe benchmark coverage and are not estimates of production
request traffic.}
\label{fig:office-coverage}
\end{figure}

\paragraph{Scale and coverage.} Figures~\ref{fig:office-composition} and~\ref{fig:office-coverage}
summarize the Office release. The first separates construction route and calibrated difficulty; the
second places task type, scenario, output family, and evaluation mechanism in one aligned row. The open release contains 50 tasks built through two routes: 30 tasks reconstructed from
task specifications and target capabilities, and 20 tasks expanded from abstracted office
workflows. Both routes produce the same release package and follow the same verification protocol.
At the broad task-family level used in this coverage view, the release contains 24 data,
spreadsheet, or structured-processing tasks; 17 document, report, or presentation tasks; and 9
workspace-automation or stateful-workflow tasks. The figure also groups tasks into six office
scenarios: data and finance analysis (16 tasks), documents
and presentation material (11), reconciliation and back-office operations (8), engineering and
tool workflows (5), stateful workflows (5), and compliance and evidence organization (5). These
groups describe benchmark coverage rather than estimate production request traffic.

Difficulty is reported in three calibrated tiers: 13 easy, 24 medium, and 13 hard tasks. Output
families use multi-label counts: 24 tasks produce spreadsheets, 20 Markdown, 15 JSON, 6 plain text,
and 5 workspace or state outputs, with smaller coverage of presentation, CSV, manifest, filesystem,
and audit-log deliverables. The release is text-first: its core tasks and evaluation do not require
OCR, a vision-language model, or pixel-level layout judgment.

\paragraph{Construction and difficulty.} Each task starts from a target capability or workflow. We
then build the agent-visible workspace and separate evaluation assets, test the evaluator on saved
submissions, calibrate difficulty, and run release checks. During execution, the agent sees only the
request and declared inputs; reference answers, expected state, rule checks, semantic rubrics, and
evaluation support files are used only after the agent finishes. Before release, saved-submission
replays check that the evaluator covers the objective requirements, provides enough evidence for the
semantic rubrics, and does not penalize valid high-quality outputs.

Difficulty comes from the solution path rather than file count alone. Common requirements include
cross-file key matching and alias resolution, temporal or state dependencies, rule priority,
conflicting or missing evidence, and consistency across multiple deliverables. A hard task may
require an agent to reconcile several sources, preserve unresolved conflicts, update both a primary
deliverable and a state record, and avoid prohibited side effects. These requirements help
distinguish model capabilities without depending on live services or undisclosed accounts.

A representative task, \texttt{hospital\_bed\_utilization}, provides a ward configuration table, an
admission log, and a bed-status policy table. The agent must compute monthly utilization by ward and
bed type and write a two-sheet workbook containing utilization detail and ward-level summaries. A
plausible-looking percentage is insufficient: the submission must resolve keys across sources,
normalize dates, apply the correct reporting period and policy denominator, preserve the requested
schema, and keep detail and summary sheets mutually consistent. The task therefore tests the
reliability of a complete file workflow rather than a single calculation.

\begin{figure}[t]
\centering
\includegraphics[width=\textwidth]{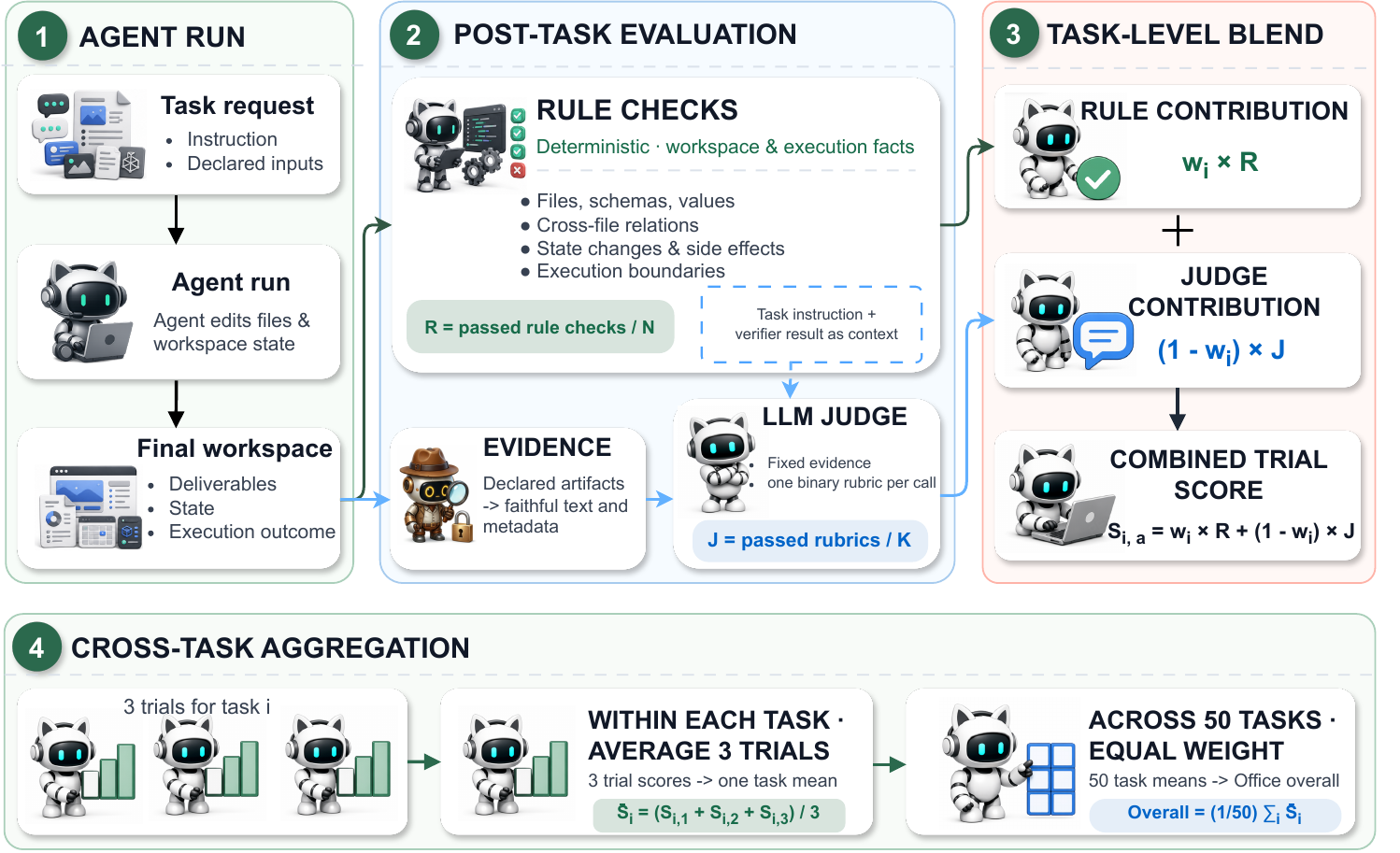}
\caption{Office task, evaluation, and scoring flow. The agent acts on the workspace and leaves a
final state (left); deterministic rule checks evaluate the verifiable workspace state while the LLM
Judge evaluates only fixed post-task evidence (center); each task combines the two scores with its
own weight, trials are averaged within each task, and task scores are macro-averaged with equal
weight (right).}
\label{fig:office-evaluation-flow}
\end{figure}

\paragraph{Scoring.} Every Office task uses two scoring components: deterministic rule checks and an
evidence-grounded LLM Judge. Rule checks are binary tests of objective requirements that can be
evaluated exactly, such as required files, schemas, values, source relations, state transitions, side
effects, and execution constraints. Each semantic rubric defines one binary quality condition that
the Judge evaluates from fixed evidence generated after the task ends, including submitted
deliverables and task-specific state or source summaries. The Judge does not inspect a live workspace
or alter recorded rule-check outcomes. All 50 tasks use both components. For selected tasks, state
differences (10 tasks), controlled environments (6), execution traces (5), or runtime boundaries
(3) provide evidence for rule checks or semantic rubrics; they are not additional scoring channels. Section~\ref{sec:harness} defines how
each task combines Rule and Judge scores, how trials are aggregated, and how unavailable Judge
results are handled. Figure~\ref{fig:office-evaluation-flow} summarizes the evaluation flow.

\subsection{Security}
\label{sec:subset-security}

The Security subset covers the security-team spectrum -- red-team discovery and safe exploitation,
malware analysis, security
operations, and agent-security assessment -- asking a sharper question than the bug-fix tasks in
Code: can an agent locate a real vulnerability and safely reproduce it in a sandboxed environment
the way a security researcher does, analyze a malware artifact or triage an alert stream the way a
malware analyst or SOC operator does, and probe a tool-using agent the way an AI red-teamer does.
Given a task, the agent must earn each step in turn, with no defect location or expected behavior
handed to it up front, and every task runs inside a sandboxed evaluation environment. What
distinguishes the subset from the rest of the suite is that it
carries no LLM judge anywhere: every task ships a deterministic scoring program that turns
agent output directly into a numeric reward, backed by a five-layer anti-cheat infrastructure that
closes off hardcoding and enumeration.

\begin{figure}[t]
\centering
\includegraphics[width=\textwidth]{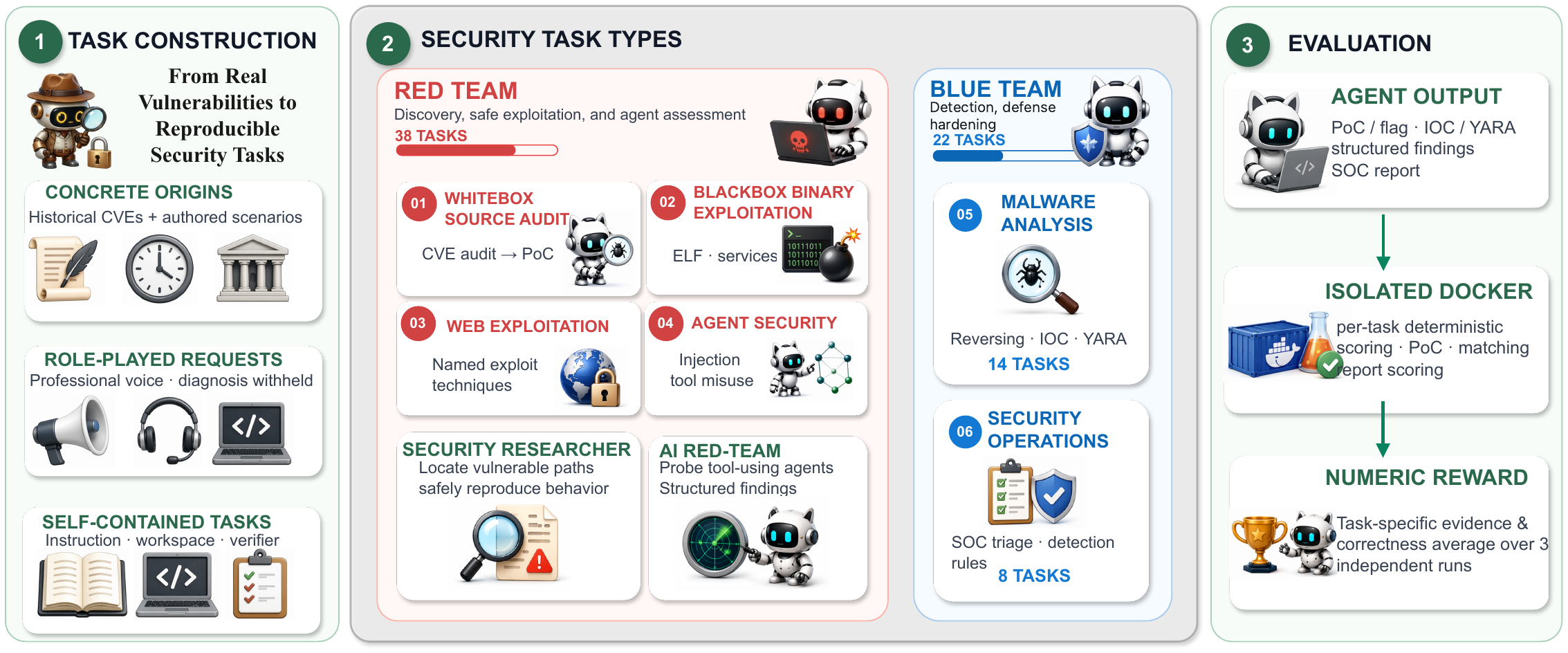}
\caption{WorkBuddy Bench Security overview. Tasks are built from real, historical vulnerabilities
and authored scenarios into reproducible, self-contained cases (left); they span six red- and
blue-team task types -- whitebox source audit, blackbox binary exploitation, web exploitation,
agent security, malware analysis, and security operations -- across 38 red-team and 22 blue-team
tasks (center); and each is scored by a per-task deterministic program inside an isolated Docker
container that emits a numeric reward (right).}
\label{fig:security-composition}
\end{figure}

The Security subset comprises \textbf{60 tasks}, spanning
six fine-grained domains rolled up into four blocks across both red-team and blue-team disciplines
(Table~\ref{tab:security-composition}, Figure~\ref{fig:security-composition}). Grouped by discipline
the suite is red-team-heavy -- 38 tasks against 22 blue-team tasks -- but still exercises the full
defend/detect loop, and difficulty skews hard by design, reflecting the balance of real security
work, where difficult cases outnumber easy ones.

\begin{table}[h]
\centering
\caption{The Security subset's composition, shown at four-block granularity (60 tasks). The
vulnerability discovery \& exploitation block subdivides into whitebox source audit, blackbox
binary exploitation, and web exploitation, giving the six fine-grained domains referenced in the
text.}
\label{tab:security-composition}
\begin{tabular}{@{}llrl@{}}
\toprule
Block & Role & Tasks & Discipline \\
\midrule
Vulnerability discovery \& exploitation & Security researcher            & 32 & Red \\
Malware analysis                        & Anti-virus engineer            & 14 & Blue \\
Security operations                     & SOC analyst / detection eng.   & 8  & Blue \\
Agent security                          & AI red-team                    & 6  & Red \\
\bottomrule
\end{tabular}

\vspace{2pt}
\footnotesize Difficulty skews hard by design.
\end{table}

Every task's deterministic scorer executes inside an isolated Docker container and writes a numeric
reward directly, so the same output re-scored twice returns the same number
(Figure~\ref{fig:security-composition}, right). Section~\ref{sec:harness} gives the per-scorer
definitions -- PoC and flag verification, IOC matching, YARA match rate under a
zero-false-positive constraint, and macro-F1/Kendall-tau report scoring.

The discovery \& exploitation block spans whitebox source-audit, blackbox binary-exploitation, and
web-exploitation tasks. The whitebox audits reproduce real, historical CVEs in widely deployed
upstream projects -- binutils, curl, nginx, vim, jq, and fluent-bit -- under a two-step
\emph{find-vuln} \(\to\) \emph{poc-verify} structure in which the second step is gated on clearing
the first. In a representative task of this kind, e.g.\ one targeting binutils, step one gives the
agent only the source tree and asks it to read the parser, trace the data flow, and locate the
vulnerable code path, scored against a threshold before the environment unlocks step two; only then
can the agent submit a proof-of-concept input, which passes only if it reproducibly triggers an
ASAN crash inside the sandboxed container -- a pacing meant to mirror a real audit-then-exploit
engagement rather than hand over the defect's location up front. The web-exploitation cases are
built around specific, named techniques (e.g., House of Apple2 and ECDSA nonce reuse) rather than
generic vulnerability classes. The six agent-security tasks probe attack surfaces specific to
tool-using AI agents -- agent-to-agent prompt injection, ReAct chain hijacking, multimodal
prompt-chain injection, tool-schema confusion, data exfiltration via a summarization tool, and
delayed-trigger attacks -- and each requires the agent to return a structured findings report with
a CVSS severity rating, mirroring the deliverable a security team would expect from a pre-launch
agent security review.

\paragraph{Anti-cheat.} To keep scores meaningful under fully automated, non-judge verification,
every task sits behind a five-layer anti-cheat infrastructure that closes off hardcoding and
enumeration along the input, code, and output axes:
\begin{itemize}\itemsep2pt
\item \emph{Banned-literal scanning} against hardcoded answers.
\item \emph{Renamed-input tests} that check whether an extractor parses structure rather than
keying off a filename.
\item \emph{Overlay/tamper tests} against trailing-data manipulation.
\item \emph{Encoding-dependence tests} that require detection rules to anchor on bytes rather than
plaintext.
\item \emph{Low-weight decoy fields} that suppress reward from blind enumeration.
\end{itemize}

Like the other subsets, Security is scored under both the CodeBuddy Code and Claude Code
harnesses in think mode, averaged over three runs; results appear in Section~\ref{sec:results}.

\newpage

\section{Evaluation Harness and Scoring}
\label{sec:harness}

A benchmark's numbers are only as trustworthy as the machinery that produces them. Tencent
WorkBuddy Bench treats that machinery as a first-class contribution rather than an
implementation detail: every task, regardless of track, ships as a self-contained task
directory and is executed inside a sandboxed container under one shared harness, and the
benchmark is released fully open -- task directories, environment images, evaluation code,
grading tests, and reference/gold solutions are all public. An external party needs no special
access to the benchmark's internal infrastructure and no bespoke evaluation path per subset: a
score can be reproduced, and any individual task re-run and audited directly, from the public
release alone. This section describes that harness, how agents connect to models under
evaluation, and the scoring rule applied per track; the per-subset sections above defer their
scoring detail here.

\paragraph{Sandboxed execution and model connectivity.} Each trial runs a task's environment
inside an isolated container; the agent sees only the task's declared workspace, and task-specific
evaluation assets are introduced into the sandbox or invoked by the evaluation pipeline only after
the agent has finished acting, so grading never leaks into the agent's context. Model and sandbox
concerns are kept deliberately separate:
the harness can reach a model backend either directly or through a local proxy that performs
protocol translation, model-name rewriting, and request logging, and it can execute the sandbox
either on a local machine or on a remote, isolated sandbox backend. When the sandbox runs
remotely, no benchmark-side proxy is ever placed inside it -- any protocol handling that would
otherwise be the proxy's job is left to the model service itself -- and the one combination of
sandbox backend and connection mode that cannot satisfy this separation is disabled outright
rather than silently falling back to another path. One access asymmetry is disclosed for
completeness: the HY (Hunyuan) endpoint used in this evaluation is served first-party by its
provider, whereas all other models are accessed through third-party serving endpoints;
third-party parameter configuration and request handling may affect metrics.

\paragraph{Harness backends.} The default execution harness is CodeBuddy Code; Claude Code, which
speaks the Anthropic protocol directly, is supported as an alternative wherever a model's own
protocol makes that route available. All four tracks are run and reported under both harnesses
side by side (dual-harness reporting), since relative rankings can shift between the two -- a
model that leads under one harness need not lead under the other (Section~\ref{sec:results}). Reasoning
mode (\texttt{think} vs.\ \texttt{nothink}) is one further configuration axis the harness records
per model, alongside the sampling hyperparameters described below; the leaderboard in
Section~\ref{sec:results} reports the think-mode configuration throughout. Across both harnesses
the protocol fixes reasoning effort to high, unifies the context window at 200k tokens with a
common auto-compaction threshold, and disables the \texttt{WebSearch} and
\texttt{AskUserQuestion} tools; each model otherwise runs with its provider-default inference
hyperparameters, as recorded below. Reported results are
tied to the specific builds of the two harnesses used in this evaluation, and metrics may shift
as harness versions evolve.

\paragraph{Scoring formalism.} Every task $t$ in a track's task set $\mathcal{T}$ yields a
verifier reward $r_t \in [0,1]$, and a model's track score is the unweighted mean
\begin{equation}
S \;=\; \frac{1}{|\mathcal{T}|} \sum_{t \in \mathcal{T}} r_t,
\end{equation}
averaged over independent runs where a track scores more than one. The per-task reward differs by
track. For Code, $r_t$ is the hidden-test pass rate of the agent's patch. For Web, the reward is
computed from a task-specific set of scored rubric items: each item returns pass/fail; let $F_t$
be the set of failed non-fatal items, each with penalty $p_{t,i}$; and any fatal failure sets the
task reward to zero:
\begin{equation}
r_t =
\begin{cases}
0, & \text{if any fatal item fails},\\
\max(0, 1 - \sum_{i \in F_t} p_{t,i}), & \text{otherwise},
\end{cases}
\end{equation}
For Office, the task reward is a task-specific blend of a deterministic Rule score and an
evidence-grounded Judge score, defined below; and for Security, the
per-task deterministic scorer combines three programmatic terms,
$r_t = w_1\,\text{artifact} + w_2\,\text{correctness} + w_3\,\text{robustness}$
(Figure~\ref{fig:security-composition}), averaged over three runs.

\paragraph{Per-track scoring.} Every task is packaged and executed the same way, but the reward
computed from it is track-specific:
\begin{itemize}
  \item \textbf{Code} -- the run-level score computed by the Harbor harness~\citep{harbor}: the
  per-run average of per-task hidden-test scores, which is the headline Code metric throughout
  this report. The verifier
  takes one of three forms -- a pytest-injected suite (22 of 80 tasks), functional boolean
  assertions needing no pytest or network access (54 of 80), or a JSON-report scorer for
  repository-understanding tasks (4 of 80); the gold patch is diagnostic only, and any satisfying
  patch scores full marks. A task-level aggregate unit-test pass rate and an LLM-judge score are
  recorded as reference values only.
  \item \textbf{Web} -- rubric scoring over 786 scored items. Rule checks cover 62 deterministic
  delivery and precheck items; LLM/VLM judges cover 676 items over text, code, structured content,
  DOM summaries, screenshots, and visual evidence; and the agent-judge covers 48 items that
  require operating the running artifact, such as workflow completion, state changes,
  persistence, and cross-state consistency. Failed items subtract configured penalties
  ($0.1/0.2/0.3$), while fatal failures set the task reward to zero. Every task must still deliver
  a runnable artifact at the declared output path, with no live-internet, external-account, or key
  access.
  \item \textbf{Office} -- two separately retained scoring channels, each composed of binary
  checks. Deterministic rule checks verify objective facts in files, structure, values, cross-file
  relations, state changes, side effects, and execution boundaries. Each semantic rubric specifies
  one binary, evidence-based quality condition evaluated by an LLM Judge after the task ends. The
  Judge receives the public task instruction, the complete set of rule-check results, and only the
  fixed evidence named by the rubric being evaluated; it does not inspect the live workspace or
  alter recorded rule-check outcomes. Each task preconfigures how the two channel scores are
  combined.
  \item \textbf{Security} -- hidden-test verification, no LLM judge: every task ships a
  \texttt{scoring.py}, run in an isolated container, writing a numeric reward directly --
  exploitation tasks verify a PoC or captured flag, malware-analysis tasks compare IOCs to ground
  truth, YARA-rule tasks check match rate under a zero-false-positive constraint, and SOC-report
  tasks score via macro-F1/Kendall-tau against a reference report. Each score averages three
  independent runs. A five-layer anti-cheat infrastructure (banned-literal scanning,
  renamed-input tests, overlay/tamper tests, encoding-dependence tests, and low-weight decoy
  fields) guards against hardcoding across the input, code, and output axes.
\end{itemize}

\paragraph{Office Rule--Judge composition.}
For model $m$ on trial $a$ of Office task $i$, let $P_{m,i,a}$ be the number of the task's $N_i$
deterministic rule checks that pass. The Rule score is
\begin{equation}
R_{m,i,a}=\frac{P_{m,i,a}}{N_i}.
\end{equation}
If task $i$ has $K_i$ semantic rubrics and rubric $k$ returns
$b_{m,i,a,k}\in\{0,1\}$, the Judge score is
\begin{equation}
J_{m,i,a}=\frac{1}{K_i}\sum_{k=1}^{K_i}b_{m,i,a,k}.
\end{equation}
The trial score uses the task's preconfigured rule weight $w_i$:
\begin{equation}
S_{m,i,a}=w_iR_{m,i,a}+(1-w_i)J_{m,i,a}.
\end{equation}
Each task fixes $w_i$ between 0.70 and 0.95; Office does not use a single global Rule weight. Let
$A_i$ be the number of available trials for task $i$. These trials are averaged first,
$\bar{S}_{m,i}=A_i^{-1}\sum_a S_{m,i,a}$. If $T$ tasks have at least one available trial, the Office
score is their equal-weight macro-average,
\begin{equation}
\operatorname{Overall}_m=\frac{1}{T}\sum_{i=1}^{T}\bar{S}_{m,i}.
\end{equation}
Rule and Judge sub-scores use the same two-level aggregation and remain available for diagnosis. A
failed evidence extraction or Judge call assigns zero only to the affected rubric; the remaining
rubrics continue. If a trial has no Judge score because the Judge input exceeds the supported length
or every Judge call fails, the evaluator retains the Rule score and error state, marks the combined
trial score as unavailable, and excludes that trial from both aggregation levels.

\paragraph{Judge-based components and scoring risk.} Three components in the suite are
model-judged rather than programmatic: Web's LLM/VLM and agent-judge rubric items, whose penalties
are configured per task; Office's LLM-judge layer for semantic-quality checks; and Code's
LLM-judge score, which is recorded as a reference value only and never enters the headline metric.
The known risk is model-judge bias -- a judge model
can systematically favor particular output styles or its own model family. The exposure is
bounded by construction: the headline metrics rest on deterministic verification for Code (hidden
tests) and Security (per-task deterministic scorer, no LLM judge); Office reduces, but does not
eliminate, LLM Judge risk by binding every binary semantic rubric to fixed post-task evidence,
retaining deterministic Rule outcomes as a separate score, and preventing Judge conclusions from
altering those outcomes; and
Web retains deterministic rule checks for delivery and precheck constraints while grounding
LLM/VLM and agent judgments in recorded evidence from the final artifact.

\paragraph{Inference hyperparameters.} A model's effective sampling behavior can be set at three
different layers -- the vendor's own server-side default, the benchmark's routing/gateway layer,
and an explicit override in the job configuration -- so the project maintains a per-model
hyperparameter record (covering fields such as temperature, top-\emph{p}, max tokens, and the
reasoning/thinking toggle) to avoid conflating the three. This bookkeeping currently spans a
schema populated for the seven evaluated models. In practice, most models are run without an
explicit sampling override, and the one parameter the benchmark deliberately fixes and reports
per model is the reasoning mode.

\paragraph{Disclosure policy.} Tencent WorkBuddy Bench is released as a fully open benchmark:
task directories, environment images, evaluation code, grading tests, and reference/gold
solutions are all made public alongside the aggregate leaderboard, following the SWE-bench-style
convention of publishing the full task set rather than an aggregate-only score. The terms ``hidden
tests'' (for Code) and ``held-out evaluation assets'' (more generally) describe solve-time
visibility, not secrecy: they are unavailable to the agent's own context during a run and are
introduced or invoked only after the agent has finished acting (see above), but they are public in
the released dataset like every other task artifact. Contamination resistance instead rests on
task freshness
at authoring time -- tasks are built from content excluded from model-pretraining corpora before
the release date -- not on withholding task content after release.

\section{Results}
\label{sec:results}

This section reports the Tencent WorkBuddy Bench leaderboard, read against the question the suite
is built to answer: how does agent capability rank across four distinct classes of real work --
Code, Web, Office, and Security -- and how robust is that ranking when the harness itself changes.
Every score is the average of three independent runs in think mode, and all four subsets are
scored under both the CodeBuddy Code (\texttt{cbc}) and Claude Code (\texttt{cc}) harnesses.
Every model is scored on every track/harness combination; one cell -- Claude Opus 4.8's Code
score under Claude Code -- comes from a modified-instruction run, marked \(\ddagger\) in
Table~\ref{tab:leaderboard} and described in its caption.

\begin{table}[htbp]
\centering
\footnotesize
\setlength{\tabcolsep}{4.5pt}
\begin{tabular}{@{}lcccccccc@{}}
\toprule
\rowcolor{wbteal!8}
 & \multicolumn{2}{c}{Code} & \multicolumn{2}{c}{Web} & \multicolumn{2}{c}{Office} & \multicolumn{2}{c}{Security} \\
\cmidrule(lr){2-3}\cmidrule(lr){4-5}\cmidrule(lr){6-7}\cmidrule(lr){8-9}
\rowcolor{wbteal!8}
Model & cbc & cc & cbc & cc & cbc & cc & cbc & cc \\
\midrule
Claude Opus 4.8   & \cellcolor{wbteal!14}\textbf{74.43} & \cellcolor{wbteal!14}\textbf{77.90}\(^{\ddagger}\) & \cellcolor{wbteal!14}\textbf{68.14} & \cellcolor{wbteal!14}\textbf{69.86} & \cellcolor{wbteal!14}\textbf{82.37} & 83.23 & 64.37 & 65.87 \\
GPT-5.5           & 72.90 & 76.63            & 61.14 & 64.86 & 81.96 & \cellcolor{wbteal!14}\textbf{86.05} & 64.39 & 77.91 \\
GLM-5.2           & 71.54 & 77.06            & 67.43 & 60.71 & 79.60 & 79.57 & \cellcolor{wbteal!14}\textbf{76.32} & \cellcolor{wbteal!14}\textbf{80.86} \\
HY-3              & 62.90 & 66.26            & 67.71 & 66.43 & 82.08 & 80.08 & 64.50 & 65.59 \\
MiniMax-M3        & 60.14 & 66.42            & 58.00 & 52.57 & 78.28 & 76.30 & 74.14 & 59.30 \\
DeepSeek-V4-Pro   & 58.92 & 64.59            & 54.57 & 51.57 & 79.11 & 78.71 & 70.04 & 58.73 \\
DeepSeek-V4-Flash & 55.73 & 61.89            & 47.29 & 50.29 & 77.47 & 77.54 & 67.11 & 53.90 \\
\bottomrule
\end{tabular}
\caption{Tencent WorkBuddy Bench leaderboard. Scores are 0--100, think mode, averaged over three
runs, under the CodeBuddy Code (\texttt{cbc}) and Claude Code (\texttt{cc}) harnesses. Bold on
shaded cells marks the best score in each column.
\(\ddagger\)Claude Opus 4.8's Code score under Claude Code comes from a modified-instruction
run: on top of the disabled \texttt{AskUserQuestion} tool (Section~\ref{sec:harness}), an
explicit do-not-ask, complete-in-one-pass instruction was added, so its setup differs slightly
from the other runs.}
\label{tab:leaderboard}
\end{table}

\textbf{Per-track leaders.} No single model tops every board. Across the eight boards in
Table~\ref{tab:leaderboard}, leadership splits three ways: Claude Opus 4.8 leads five -- Code
under both harnesses (74.43 under \texttt{cbc}; 77.90 under \texttt{cc}, from the
modified-instruction run noted \(\ddagger\) in the table caption), Web under both harnesses
(68.14 under \texttt{cbc}, 69.86 under \texttt{cc}), and Office under \texttt{cbc} (82.37),
where HY-3 (82.08) is the Office runner-up just below it; GLM-5.2 leads two -- Security under
both harnesses (76.32 under \texttt{cbc}, 80.86 under \texttt{cc}); and GPT-5.5 leads one,
Office under \texttt{cc} (86.05). That an open-weight model, GLM-5.2, tops both Security boards
outright is itself a finding: on this suite the gap between open-weight and closed frontier
models is board-dependent rather than uniform.

\textbf{Harness sensitivity.} With all four tracks scored under both harnesses, the harness is
visibly not a neutral measurement instrument -- and the four tracks are affected to very
different degrees. Code shifts the most uniformly: the rank order of models on Code
differs between the two harnesses --- GPT-5.5 sits ahead of GLM-5.2 under
\texttt{cbc} (72.90 vs.\ 71.54) but behind it under \texttt{cc} (76.63 vs.\ 77.06) ---
and Claude Opus 4.8's modified-instruction \texttt{cc} run likewise sits higher than its
\texttt{cbc} score.\footnote{Harness--model integration details also matter within a
single harness: a diagnostic rerun of HY-3 on the Code subset with cross-turn reasoning passback
enabled -- thinking content passed back to the backend across turns -- scored 66.72 under
CodeBuddy Code ($+3.82$ over the leaderboard configuration) and 68.18 under Claude Code
($+1.92$), under the same three-run protocol. The leaderboard reports the standard
configuration.} Web is more mixed under Claude Code:
four of seven dual-scored models drop, by margins from $-1.28$ (HY-3) to $-6.72$
(GLM-5.2), while three rise -- Claude Opus 4.8 ($+1.72$), DeepSeek-V4-Flash ($+3.00$), and
GPT-5.5 ($+3.72$); the signed mean shift is $-1.14$, and Claude Opus 4.8 leads Web under both
harnesses.
Office moves least: five of the seven dual-scored models shift by under two points (median absolute
shift 0.86), the exceptions being GPT-5.5 ($+4.09$) and HY-3 ($-2.00$).
Security shows the largest reordering between harnesses: the mean absolute shift
across the seven dual-scored models is 8.6 points. GLM-5.2 leads Security under both
harnesses, but below it the board reorders substantially: GPT-5.5 climbs from sixth under
\texttt{cbc} to second under \texttt{cc}, while MiniMax-M3
falls from second to fifth.

\textbf{Refusals on Security.} A small number of Security runs ended in task-level refusals on
security-flavored requests. Across the three runs, Claude Opus 4.8 recorded 13 refusals under
Claude Code (none under CodeBuddy Code), GPT-5.5 recorded 2 under CodeBuddy Code,
and all other models recorded none. These counts
are reported for context; the leaderboard scores average over all runs as executed.

\textbf{Coding versus Data \& Algo gap.} The Code subset's own category taxonomy separates coding
proper from data- and algorithm-style work, and the gap between them is systematic rather than
incidental: a per-category breakdown (not shown in Table~\ref{tab:leaderboard}, drawn from the
subset's internal per-category results) finds that model/harness configurations almost uniformly
score higher on Data\,\&\,Algorithm tasks than on Coding tasks, at an average of roughly 74\%
versus roughly 65\%. The reading offered alongside that breakdown is that data- and algorithm-style tasks are
rarely hard \emph{as code} -- they are hard in business or data semantics -- whereas precisely
fixing a real repository under an existing contract is the more discriminating skill. Table~\ref{tab:leaderboard}
is consistent with this: in every model/harness configuration scored on both tracks, the Code
score sits below the same configuration's Office score, often by ten points or more, whereas the
Code score sits above the same configuration's Web score in all but one case -- HY-3 is the sole
exception, scoring higher on Web than on Code under both harnesses, clearly under \texttt{cbc}
(67.71 vs.\ 62.90) and marginally under \texttt{cc} (66.43 vs.\ 66.26).

\textbf{Which Code categories are hardest.} A per-category breakdown of the Code subset (mean
reward averaged over all valid configurations) makes the same point at finer grain. The two
hardest categories are \texttt{bug\_fix} (mean 0.47) and \texttt{api\_contract} (mean 0.47) --
real-repository regression fixing and precise, contract-honoring interface work -- while the
easiest are \texttt{feature\_pipeline} (0.94) and \texttt{testing} (0.88), which are well-specified
synthetic pipelines and test-writing tasks. Several product/analytics categories show an
exceptionally wide model spread (\texttt{product\_analytics} ranges from 0.08 to 1.00 across
models), a signature of tasks where the score turns on whether the model correctly reads the
business intent rather than on whether its code runs.

\textbf{Why \texttt{bug\_fix} is the hardest category.} These tasks are real upstream regressions
posed colloquially, with the root cause withheld. Solving one means locating an intermittent,
context-dependent fault from a one-sentence symptom description -- pure repository understanding
with no algorithmic difficulty -- and then patching it minimally without breaking the surrounding
contract. The low mean (0.47) shows that current models still struggle at precisely this
SWE-bench-style skill of grounding a vague report in the right lines of a large codebase. A
related, distinct failure is \emph{semantically correct but contract-mismatched} code: the model
implements the right behavior under the wrong function name, parameter shape, or output format, so
a functional verifier still fails it. This is most acute on \texttt{api\_contract}, where a task
must preserve a precise set of fields under an existing contract and a single dropped field zeroes
the checks that depend on it.

\textbf{Two representative failure modes.} Two Code bad cases illustrate the dominant ways points
are lost. In an OpenAPI contract task, a model produces behaviorally reasonable output but drops
one or two required fields (for example \texttt{deprecated} or \texttt{examples}) or leaves a path
parameter's \texttt{required} flag to its default; because the verifier runs a dozen per-field
boolean checks, each omission zeroes the checks that depend on it and drags the overall score down
sharply -- the functionality is not wrong so much as the contract implied by the tests is not
matched. In a product-analytics task, the colloquial ask is to compute per-group conversion and
revenue while ``not counting purchases that happen long afterward'' -- an implicit
conversion-attribution window. High-scoring models filter the late purchases by that window;
low-scoring ones ignore the constraint and count every purchase, over-estimating conversion, even
though the code runs cleanly either way. The gap comes entirely from whether the business rule was
understood, which is why the model spread on this category runs nearly the full range.

\textbf{Web capability slices.} Web slice results point to a consistent pattern: visual design and
analytical reporting are the strongest categories, with code testing and page implementation next,
while page interaction and data-visualization semantics expose the most failures. The
interaction/state axis tells a complementary story: noninteractive and lightly interactive
artifacts score well above stateful ones -- single-flow state changes, multi-step workflows, and
persistence, offline, and cross-state behavior are the hardest slices in the subset.

The failure pattern is less about rendering a visible page than about closing a front-end
engineering loop. Models often produce a plausible UI but lose consistency among state source,
display, persistence, and final payload -- the interactive and stateful slices are precisely where
scores are lowest -- or produce charts and data-visualization outputs without a clear
source-to-output evidence trail. By evaluation signal, LLM/VLM items account for most checks and most failed items; rule
failures mostly reflect delivery, precheck, format, or executable-test contracts, while
agent-judge failures correspond to workflow or state breakage in the running artifact.

\begin{figure}[htbp]
\centering
\includegraphics[width=\textwidth]{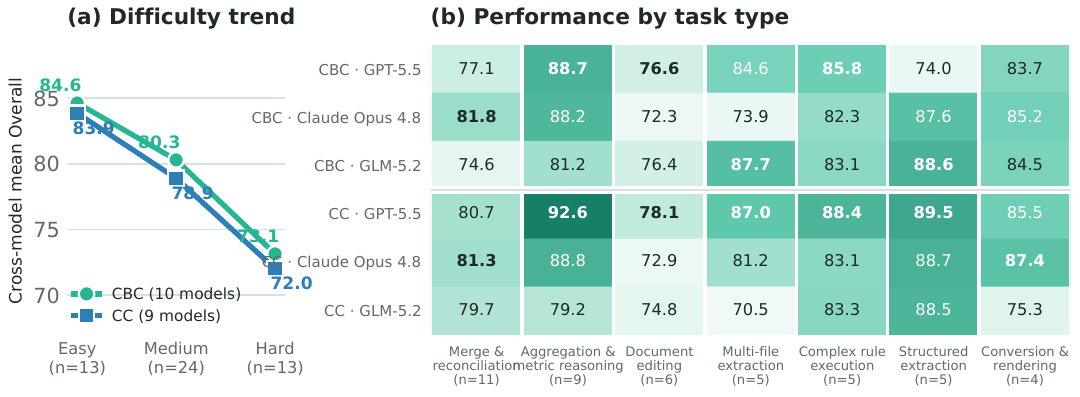}
\caption{Office results by difficulty and task type. (a) Mean Office score by calibrated difficulty
within each harness, averaged over all models with scored runs in that panel. (b) Equal-weight task
average for the seven diagnostic task types represented by at least four tasks, shown for the three
models discussed in the text. Scores use a 0--100 scale. Bold marks the best score in the full model
set for each task type and harness, so a displayed three-model group may contain no bold value.
HY-3's per-task slices in this figure predate its updated Office run, so its contribution to the
panel means reflects the earlier scoring.}
\label{fig:office-results}
\end{figure}

\textbf{Office performance by difficulty and task type.} Figure~\ref{fig:office-results} shows how
Office performance varies by difficulty and by seven diagnostic task types. Within each harness,
cross-model mean scores decline from easy to medium to hard tasks (84.6/80.3/73.1 under
\texttt{cbc} and 83.9/78.9/72.0 under \texttt{cc}). Model strengths also differ by task type:
Claude Opus 4.8 leads multi-source merge and reconciliation under both harnesses, whereas GPT-5.5
leads five of the seven types shown under \texttt{cc}, including aggregation and metric reasoning,
complex rule execution, and structured extraction. This view also reveals differences hidden by
aggregate scores: GLM-5.2 is nearly unchanged overall across harnesses (79.60 vs.\ 79.57), yet its
multi-file extraction score falls from 87.7 to 70.5 while complex rule execution remains nearly
unchanged (83.1 vs.\ 83.3).

\textbf{Common failure patterns in reviewed Office submissions.} We inspected the deliverables and
Rule/Judge evidence for selected low-scoring tasks. The same problems recurred: related deliverables
were inconsistent, records were not clearly tied to their source or current state, structured files
could not be parsed, or agents submitted artifacts without validating them. Because the Office
verifier checks files, cross-file relations, state, output contracts, and evidence-grounded semantic
requirements, these incomplete workflows lose points even when one deliverable looks plausible.

\subsection{Token and Turn Efficiency}
\label{sec:results-efficiency}

Table~\ref{tab:efficiency} reports per-run averages of assistant turns, output tokens, and input
tokens on the Code subset. Turns are counted as unique assistant messages, including subagent
activity alongside the main agent. Output tokens are comparable across the two harnesses; input
tokens are reported cache-inclusive -- they count cached context reads as well as fresh input --
and are \emph{not} comparable across harnesses, because the two harnesses manage context and
caching under different conventions, so input figures should only be read within a harness column.
Output-token counts should also not be read as a cross-model efficiency metric:
different models use different tokenizers, so a token is not a constant unit of work across
models, and the comparisons below are illustrative rather than a rigorous cross-model efficiency
ranking.

\begin{table}[htbp]
\centering
\footnotesize
\setlength{\tabcolsep}{5pt}
\begin{tabular}{@{}lcccccc@{}}
\toprule
\rowcolor{wbteal!8}
 & \multicolumn{3}{c}{CodeBuddy Code (cbc)} & \multicolumn{3}{c}{Claude Code (cc)} \\
\cmidrule(lr){2-4}\cmidrule(lr){5-7}
\rowcolor{wbteal!8}
Model & Avg turns & Output (k) & Input (k) & Avg turns & Output (k) & Input (k) \\
\midrule
Claude Opus 4.8   & 29.51 & 22.3 & 928.7  & 13.2\(^{\ddagger}\) & 4.7\(^{\ddagger}\) & 646.5\(^{\ddagger}\) \\
GPT-5.5           & 26.92 & 6.9  & 753.2  & 30.44 & 8.7  & 696.5  \\
GLM-5.2           & 33.06 & 12.3 & 861.4  & 33.73 & 22.0 & 1243.3 \\
HY-3              & 26.02 & 9.3  & 586.8  & 18.07 & 13.9 & 659.2  \\
MiniMax-M3        & 28.89 & 8.5  & 1021.4 & 33.90 & 10.6 & 1308.3 \\
DeepSeek-V4-Pro   & 44.01 & 10.2 & 800.0  & 24.20 & 23.7 & 642.3  \\
DeepSeek-V4-Flash & 40.30 & 9.8  & 700.5  & 23.12 & 28.6 & 771.4  \\
\bottomrule
\end{tabular}
\caption{Per-run averages on the Code subset, by harness: turns (unique assistant messages,
including subagent activity), output tokens, and cache-inclusive input tokens, in thousands.
Output tokens are comparable across harnesses; input tokens are not, because the two harnesses
manage context and caching under different conventions.
\(\ddagger\)Claude Opus 4.8's Claude Code figures come from the modified-instruction run
described in Table~\ref{tab:leaderboard}.}
\label{tab:efficiency}
\end{table}

Three observations. First, GPT-5.5 posts top-tier scores on a minimal output budget: its 6.9k
tokens per run under \texttt{cbc} is the lowest output budget on that harness, and its 8.7k under
\texttt{cc} is the lowest among the standard-protocol runs -- against a field that mostly spends
8--29k; the smallest \texttt{cc} output overall belongs to Claude Opus 4.8's
modified-instruction run (4.7k). Second, spend and rank are not aligned: DeepSeek-V4-Flash emits roughly
3.3$\times$ GPT-5.5's output under \texttt{cc} (28.6k vs.\ 8.7k) while scoring 14.74 points lower
(61.89 vs.\ 76.63), and GLM-5.2's 77.06 under \texttt{cc} costs 22.0k output tokens against GPT-5.5's
8.7k for a 0.43-point margin -- while Claude Opus 4.8 pairs the highest
\texttt{cbc} output (22.3k) with the \texttt{cbc} lead. Third, among standard-protocol runs turn
counts vary roughly
2.4$\times$ across configurations (18.07 to 44.01 per run), and neither extreme aligns with
score: the leanest standard configuration is HY-3 under \texttt{cc} at 18.07 turns for a mid-board 66.26,
while the two largest turn counts (DeepSeek-V4-Pro at 44.01 and DeepSeek-V4-Flash at 40.30,
both under \texttt{cbc}) belong to the two lowest-scoring \texttt{cbc} configurations. Claude
Opus 4.8's modified-instruction \texttt{cc} run sits below the standard range at 13.2 turns and
4.7k output tokens per run.

\paragraph{Across subsets.} The Code budget profile is not universal. Office runs are the
leanest -- 16--42 turns and 10--30k output tokens per run across models and harnesses -- and Web
sits in the middle (13--39 turns), while Security is the heaviest track by a wide
margin, at 30--89 turns per run.\footnote{Security turn and token statistics still use the
earlier turn-counting convention and have not yet been recomputed as unique assistant messages;
their turn counts are therefore not directly comparable to the Code figures in
Table~\ref{tab:efficiency}.} The Security extreme is stark: MiniMax-M3 under \texttt{cbc}
averages 88.8 turns and roughly 11.1M cache-inclusive input tokens per run for its 74.14. Across all four tracks, GPT-5.5 is consistently the leanest high
scorer: under \texttt{cbc} it posts the smallest output budget of any model on every track --
6.9k output tokens per run on Code, 13.5k on Web, 10.2k on Office, and 7.5k on Security.

\section{Related Work}
\label{sec:related-work}

We first position Tencent WorkBuddy Bench against existing agent benchmarks in the code, web,
office, and security domains; the suite's own task-construction methodology, subset design, and
evaluation harness are detailed in the sections that follow. These are design-time comparisons drawn from the benchmark's
own qualitative analysis of task design, not a head-to-head measured evaluation of agents across
these suites.

\textbf{Code.} The Code subset occupies a similar problem space to the SWE-bench
family~\citep{jimenez2024swebench, openai2024sweverified} and to library-from-scratch benchmarks
such as Commit0~\citep{zhao2024commit0}, but differs in instruction style and role diversity. Where
SWE-bench and SWE-bench Verified supply a detailed GitHub issue -- and Commit0 a test-driven
specification to implement against -- Code tasks are authored as short, colloquial requests,
closer to how a teammate phrases an ask than to a filed issue, deliberately leaving implementation
detail underspecified; and Code spans five requester roles (developer, algorithm engineer, product
manager, QA, ops) across 18 categories beyond bug fixing, rather than a single issue-resolution
framing. Contamination resistance is pursued differently across the family: LiveCodeBench~\citep{jain2024livecodebench}
relies on problems released after model training cutoffs, whereas Code relies on freshly authored
task directories -- real upstream commits, clean-room reimplementations, and synthetic
workspaces -- built and held back from publication until the benchmark's release, at which point
prompts, hidden tests, and gold patches are published in full alongside it.
RepoBench~\citep{liu2024repobench} and Aider Polyglot~\citep{aiderpolyglot} target narrower slices
of the same space (repository-level completion and templated multi-language exercises), and
Terminal-Bench~\citep{terminalbench2024} evaluates general terminal-agent competence rather than
repository-scoped code changes.

\textbf{End-to-end and production coding-agent benchmarks.}
Vibe Code Bench~\citep{tran2026vibecodebench} evaluates zero-to-one web application development
from text specifications through browser-agent workflow tests over deployed applications, making
it a close reference point for runnable front-end deliverables. CursorBench~\citep{cursor2026cursorbench}
pursues a related realism goal by a different route: it traces committed code back to the original
agent request from authentic production sessions, so its task distribution is anchored to how one
vendor's users actually work rather than to curated issue text. These choices make both benchmarks
important reference points, but they leave different gaps for our setting. Vibe Code Bench focuses
on from-scratch application construction, while WorkBuddy Web also covers modification, review,
front-end project tests, analysis, and conversion. CursorBench is closed-source, so its task set,
category distribution, and any selection bias toward the vendor's own agent cannot be independently
audited, and its representativeness cannot be confirmed to extend beyond that vendor's user base.
Tencent WorkBuddy Bench pursues realism through a distribution-informed route and then releases the
result fully open: task categories, shapes, and intents are checked against real usage
(Section~\ref{sec:the-benchmark}), tasks are reverse-engineered from real artifacts and curated or
synthesized to match that distribution rather than released as raw production sessions, and the
resulting task directories, environment images, evaluation code, grading tests, and reference
solutions are all publicly released and independently auditable.

\textbf{Web.} Design2Code~\citep{si2024design2code} and Interaction2Code~\citep{xiao2024interaction2code}
evaluate static and lightly interactive page reproduction from a reference design;
FrontendBench~\citep{liu2024frontendbench} extends automatic judging to a broader set of
front-end generation tasks; WebArena~\citep{zhou2024webarena} and VisualWebArena~\citep{koh2024visualwebarena}
instead evaluate an agent operating an existing browser environment rather than producing a
runnable artifact from scratch. Each is strong on one or two axes -- static reproduction,
interactive generation, browser-agent operation, or code-maintenance realism -- but none combines
page/UI work, data and chart artifacts, front-end project documents, tests, and analyses,
non-scratch lifecycle coverage, runtime interaction/state checks, and rule, LLM/VLM, and agent
judging in one evaluation. Table~\ref{tab:webmatrix} keeps those axes
separate rather than reporting task counts, since published benchmark scales are not directly
comparable across webpage, interaction, issue, and application-specification units. The comparison
is qualitative and self-reported from each benchmark's own published description, not a measured
evaluation.

\begin{table}[htbp]
\centering
\scriptsize
\setlength{\tabcolsep}{2.7pt}
\resizebox{\textwidth}{!}{%
\begin{tabular}{@{}lcccccccccccc@{}}
\toprule
\rowcolor{wbteal!8}
 & \multicolumn{4}{c}{Work surface} & \multicolumn{3}{c}{Lifecycle} &
\multicolumn{2}{c}{Runtime evidence} & \multicolumn{3}{c}{Oracle} \\
\cmidrule(lr){2-5}\cmidrule(lr){6-8}\cmidrule(lr){9-10}\cmidrule(l){11-13}
\rowcolor{wbteal!8}
Benchmark & UI & App & Data & Doc/test & Scratch & Fix/ext. & Review/convert &
Action & State & Rule & VLM & Agent \\
\midrule
Vibe Code Bench v1.1 & \(\bullet\) & \(\bullet\) & \(\circ\) &  & \(\bullet\) &  &  &
\(\bullet\) & \(\circ\) & \(\bullet\) &  & \(\circ\) \\
CursorBench 3.1      & \(\circ\) &  &  & \(\circ\) & \(\circ\) & \(\bullet\) & \(\bullet\) &
 &  & \(\circ\) &  & \(\bullet\) \\
Design2Code          & \(\bullet\) &  &  &  & \(\bullet\) &  &  &
 &  & \(\circ\) & \(\bullet\) &  \\
Interaction2Code     & \(\bullet\) & \(\bullet\) &  &  & \(\bullet\) &  &  &
\(\bullet\) & \(\circ\) & \(\bullet\) & \(\circ\) &  \\
FrontendBench        & \(\bullet\) & \(\circ\) &  &  & \(\bullet\) &  &  &
\(\bullet\) & \(\circ\) & \(\bullet\) &  &  \\
WebArena             &  & \(\bullet\) &  &  &  &  &  &
\(\bullet\) & \(\bullet\) & \(\bullet\) &  &  \\
VisualWebArena       &  & \(\bullet\) &  &  &  &  &  &
\(\bullet\) & \(\bullet\) & \(\bullet\) & \(\circ\) &  \\
\rowcolor{wbteal!10}
\textbf{WorkBuddy Web} & \(\bullet\) & \(\bullet\) & \(\bullet\) & \(\bullet\) &
\(\bullet\) & \(\bullet\) & \(\bullet\) & \(\bullet\) & \(\bullet\) &
\(\bullet\) & \(\bullet\) & \(\bullet\) \\
\bottomrule
\end{tabular}
}%
\caption{Web benchmark capability matrix. \(\bullet\)~= full coverage, \(\circ\)~= partial
coverage, blank~= not covered or not applicable. Axes are task-design and verification coverage,
not a head-to-head measured evaluation of agent performance; model-judge bias for VLM and
agent-judge layers is discussed separately in Section~\ref{sec:limitations}.}
\label{tab:webmatrix}
\end{table}

\textbf{Office.} Recent benchmarks cover complementary parts of office-agent work.
Workspace-Bench 1.0~\citep{tang2026workspacebench} evaluates tasks with large-scale heterogeneous
file dependencies using fine-grained rubrics across multiple agent harnesses.
ClawsBench~\citep{li2026clawsbench} evaluates capability and safety in snapshot-restored simulations
of Gmail, Calendar, Docs, Drive, and Slack. OdysseyBench~\citep{wang2025odysseybench} targets
long-horizon, multi-application workflows over extended interaction histories, while SpreadsheetBench
2~\citep{zhu2026spreadsheetbench2} probes end-to-end construction, repair, and visualization in
complex multi-sheet workbooks.

ClawsBench and OdysseyBench emphasize interaction across multiple applications, SpreadsheetBench 2
focuses on workbook workflows, and Workspace-Bench addresses heterogeneous file dependencies.
WorkBuddyBench-Office, the suite's Office subset, focuses on complete handoffs in local workspaces
containing multiple file formats. Agents must carry source information into deliverables, keep
related files and state consistent, and respect execution constraints under CodeBuddy Code or
Claude Code. Deterministic rule checks verify files, cross-file relations, state changes, side
effects, and execution constraints, while an evidence-grounded LLM Judge scores binary semantic
rubrics from fixed post-task evidence. Each task sets its own Rule/Judge weight. These comparisons
concern task and verification design; the benchmarks use different task units, environments, and
scoring schemes.

\textbf{Security.} Existing security-agent benchmarks each cover a slice of the red-team side:
Cybench~\citep{zhang2025cybench} and NYU CTF Bench~\citep{shao2024nyuctf} score
professional-level and competition CTF challenges, InterCode-CTF~\citep{yang2023intercode} casts
CTF solving as interactive coding with execution feedback,
CVE-Bench~\citep{zhu2025cvebench} measures autonomous exploitation of real-world web-application
CVEs, and Meta's CyberSecEval series~\citep{bhatt2024cyberseceval2, wan2024cyberseceval3}
assesses cybersecurity risks and capabilities of the models themselves, from insecure code
suggestions to offensive-operation assistance. The Security subset differs on two axes: coverage
and scoring. Its 60 tasks span red- and blue-team work in a single suite -- vulnerability
discovery and safe exploitation, malware analysis, security operations, and agent security --
rather than CTF or exploitation alone, and every task is scored by a deterministic per-task
\texttt{scoring.py} behind five anti-cheat layers, with no LLM judge anywhere. Its whitebox
discovery tasks are anchored to real, historical CVEs in widely deployed upstream projects,
rebuilt as rewritten, sandboxed environments kept out of public training corpora for
contamination resistance.

\textbf{What the suite adds.} Breadth and framing differentiate the suite, not a new task type.
Code pairs repository-scale tasks with five requester roles and an 18-category taxonomy on
colloquial asks, not filed-issue prompts, resistant to searchable-prompt and leaked-answer
contamination by construction through freshly authored task directories held back until release.
Web unifies the task-type, lifecycle-mode, interaction/state, and judge axes above in one family,
combining rule checks, LLM/VLM judgment, and agent-judge verification over running front-end
artifacts. Office treats mixed-format, multi-artifact workflows as complete handoffs, measuring
machine-checkable workspace state and semantic deliverable quality through separately retained Rule
and Judge channels. Security covers the red- and blue-team spectrum
under fully deterministic per-task scoring, in a domain where public benchmarks remain scarce.
Resistance to searchable prompts and leaked tests or answers rests on freshly authored task
directories held back until release and by-construction design; distribution-informed construction
rests on distribution-matched tasks, released fully open -- task directories, environment images,
evaluation code, grading tests, and reference solutions all public -- and so independently
auditable in full where closed vendors cannot be. These remain the benchmark's own design-time
positioning claims, not a measured comparison of agent performance across the compared suites.

\section{Limitations and Conclusion}
\label{sec:limitations}

This section discusses current limitations of Tencent WorkBuddy Bench as described in this report,
and the near-term work planned to address them.

\begin{itemize}
\item \textbf{One leaderboard cell uses a modified instruction setup.} All seven models are
scored under both harnesses on all four tracks. The one comparability caveat is Claude Opus
4.8's Code score under Claude Code: as noted in Table~\ref{tab:leaderboard}, on top of the
disabled \texttt{AskUserQuestion} tool, an explicit do-not-ask, complete-in-one-pass instruction
was added for that run, so its setup differs slightly from the other runs, and its score is
reported alongside -- but not folded into -- the Code harness-shift aggregate in
Section~\ref{sec:results}.

\item \textbf{Single-language emphasis in Code.} The Code subset's open release is dominated by
Python tasks; cross-language coverage is limited to a small number of tasks that port target
behavior from JavaScript, TypeScript, or Rust projects into Python. Findings in this report about
coding difficulty, the coding-versus-data/algorithm gap, and harness divergence should not be
assumed to generalize to other programming languages or ecosystems without further evaluation.

\item \textbf{Open release creates post-release contamination exposure.} Tencent WorkBuddy Bench
is released fully open -- task directories, environment images, evaluation code, grading tests,
and reference solutions are all public, so that external readers can independently re-run and
audit individual task outcomes, not just reproduce the pipeline described earlier in this report.
The cost of that openness is that published task content is, from the moment of release, exposed
to being scraped into future model training data, which can erode contamination resistance over
time. This is mitigated, not eliminated, by dataset versioning -- future revisions can retire or
replace tasks that show contamination symptoms -- rather than by withholding task content.

\item \textbf{Judge-based components carry model-judge bias.} Web scoring combines rule-based
checks with LLM/VLM and agent-judge rubric items over evidence from the running front-end
artifact; Office combines deterministic rule checks with an LLM Judge over fixed post-task
evidence; and Code additionally computes a diagnostic, weighted LLM-judge score across several
dimensions that is not counted in the headline metric. Office retains rule-check outcomes
independently, so the Judge cannot alter them, but its semantic-rubric scores remain subject to
model-judge bias. More generally, model judges may favor response styles they find familiar or
legible, independent of task correctness -- a risk this report has not separately quantified.

\item \textbf{Scores are tied to specific serving and harness conditions.} The HY (Hunyuan)
endpoint used in this evaluation is served first-party by its provider, while all other models
are accessed through third-party serving endpoints, whose parameter configuration and request
handling may affect metrics. Likewise, results are tied to the specific builds of the two
harnesses used in this evaluation, and metrics may shift as harness versions evolve
(Section~\ref{sec:harness}).

\item \textbf{Office is text-first.} The current Office release covers local, mixed-format workflows
but does not require OCR, vision-language models, pixel-level layout judgment, or native desktop-GUI
interaction. Office results therefore apply to file handling, state updates, and evidence-based
workflow completion, not to visual perception or GUI operation.
\end{itemize}

Near-term work focuses on calibration: continuing Web rubric calibration, and, where feasible,
bringing the one modified-instruction configuration (Claude Opus 4.8 on Code under Claude Code)
under the standard instruction protocol.

This report has described Tencent WorkBuddy Bench as released: four subsets sharing one
task-directory format, one admission protocol, and one execution harness, together with its
leaderboard and the limitations stated above. These limitations and scope boundaries are the ones
we consider material enough to state explicitly, not an exhaustive list. The suite is released fully
open -- task directories, environment
images, evaluation code, grading tests, and reference solutions are all public for offline
third-party testing and audit -- and the near-term trajectory beyond this revision is a public
leaderboard with expanding model coverage across both harnesses.

\section*{Contributors}
\label{sec:contributors}

Siqi Cai\textsuperscript{1*}, Shaopeng Chen\textsuperscript{4*}, Xiang Fei\textsuperscript{1*},
Yong Mao\textsuperscript{1*}, Zihan Xu\textsuperscript{1*}, Zhiheng Lyu\textsuperscript{3*},
Zhijian Shao\textsuperscript{2*}, Yuchen Shi\textsuperscript{1}, Shuwen Zhang\textsuperscript{1},
Chaofan Qiu\textsuperscript{1}, Linjie Che\textsuperscript{3}, Xiaoxi Zhao\textsuperscript{3},
Feng Wu\textsuperscript{3}, Kai Zhang\textsuperscript{3}, Chaofan Zhu\textsuperscript{3}, Yubin Qi\textsuperscript{3},
Xiaoyun Liang\textsuperscript{3}, Peijie Dong\textsuperscript{3}, Yunhao Zhang\textsuperscript{3}, Yuanjie Zhu,
Ling Jiang\textsuperscript{2}, Xianjun Zhang\textsuperscript{2}, Zhehang Chu\textsuperscript{2}, Anyuan Sang\textsuperscript{2},
Zhen Feng\textsuperscript{2}, Sen Nie\textsuperscript{2}, Shi Wu\textsuperscript{2},
Yuanzhen Xu\textsuperscript{4}, Xin Li\textsuperscript{4}, Ning Yang\textsuperscript{4},
Zhiqiang Dong\textsuperscript{4}, Hande Dong\textsuperscript{3}, Qiang Lin\textsuperscript{3},
Yi Liu\textsuperscript{3}, Yunsheng Wu\textsuperscript{1},
Ke Li\textsuperscript{1\textdagger}, Xing Sun\textsuperscript{1}

\medskip

\noindent\textsuperscript{1}Youtu Lab \(\cdot\) \textsuperscript{2}Keen Security Lab \(\cdot\)
\textsuperscript{3}Workbuddy \(\cdot\) \textsuperscript{4}Yunding Security Lab

\medskip

{\small
\noindent *These authors contributed equally to this work. The author order was determined
alphabetically.\\
\textdagger Project Lead.
}

\microtypesetup{expansion=false}
\bibliography{refs}

\end{document}